\PassOptionsToPackage{numbers,compress}{natbib}
\documentclass{article}
\usepackage{iclr2027_conference,times}
\usepackage{graphicx}
\usepackage{float}
\usepackage{booktabs}
\usepackage{multirow}
\usepackage[table]{xcolor}
\usepackage{threeparttable}
\usepackage[most]{tcolorbox}
\usepackage{tikz}
\usepackage{amsmath,amssymb}
\usepackage{booktabs,array,tabularx,threeparttable}
\usetikzlibrary{tikzmark}

\usepackage{xcolor}

\definecolor{revision}{RGB}{220,50,47}
\definecolor{todo}{RGB}{0,100,200}

\definecolor{jrcolor}{RGB}{225, 100, 150}

\usepackage{amsmath,amsfonts,bm}

\def\eqref#1{equation~\ref{#1}}

\def\1{\bm{1}}

\DeclareMathAlphabet{\mathsfit}{\encodingdefault}{\sfdefault}{m}{sl}
\SetMathAlphabet{\mathsfit}{bold}{\encodingdefault}{\sfdefault}{bx}{n}

\input{appendix_genrevisit/appendix_dependencies}
\usepackage{hyperref}
\input{appendix_genrevisit/appendix_styles}
\AtBeginEnvironment{table}{\setlength{\abovecaptionskip}{0pt}\setlength{\belowcaptionskip}{4pt}}
\usepackage[all]{hypcap}
\hypersetup{hidelinks}
\usepackage{url}
\setcitestyle{numbers,square,comma}

\title{Look Back to Move Forward: Learning to Revisit for Agentic Compositional Image Generation}

\usepackage{fontawesome5}
\iclrfinalcopy
\author{\parbox[t]{\dimexpr\textwidth-2\tabcolsep\relax}{\raggedright\bfseries
\mbox{Zixu Wang\textsuperscript{1,*}}\hspace{0.5em}
\mbox{Xiaoran Zhao\textsuperscript{1,2,*}}\hspace{0.5em}
\mbox{Yahui Liu\textsuperscript{3,\ensuremath{\dagger}}}\hspace{0.5em}
\mbox{Feng-ao Wang\textsuperscript{4,\ensuremath{\dagger}}}\hspace{0.5em}
\mbox{Junhao Xiao\textsuperscript{5}}\hspace{0.5em}
\mbox{You Shu\textsuperscript{1,2}}\hspace{0.5em}
\mbox{Bing Liu\textsuperscript{6}}\hspace{0.5em}
\mbox{Yiting Liu\textsuperscript{7}}\hspace{0.5em}
\mbox{Songlei Jian\textsuperscript{1}}\hspace{0.5em}
\mbox{Zhiliang Tian\textsuperscript{1,2}}\hspace{0.5em}
\mbox{Dongsheng Li\textsuperscript{1,2}}
\endgraf\vspace{5pt}\normalfont\small
\mbox{\textsuperscript{1}NUDT}\hspace{0.5em}
\mbox{\textsuperscript{2}PDL Lab}\hspace{0.5em}
\mbox{\textsuperscript{3}Xiaohongshu}\hspace{0.5em}
\mbox{\textsuperscript{4}HNU}\hspace{0.5em}
\mbox{\textsuperscript{5}FDU}\hspace{0.5em}
\mbox{\textsuperscript{6}SCUT}\hspace{0.5em}
\mbox{\textsuperscript{7}SYSU}
\endgraf}}
\renewcommand{\footnoterule}{\kern-3pt\hrule width\textwidth\kern2.6pt}
\makeatletter
\gdef\@thanks{\begingroup
\long\def\@makefntext##1{\noindent##1}
\renewcommand{\thefootnote}{}
\footnotetext{\textsuperscript{*}Equal contribution.\quad
\textsuperscript{\ensuremath{\dagger}}Corresponding authors.\hfill
\href{https://github.com/Flipped-zx/GenRevisit}{\textcolor{black}{\faGithub\enspace\textnormal{https://github.com/Flipped-zx/GenRevisit}}}}
\endgroup}
\makeatother

\makeatletter
\patchcmd{\@maketitle}{Published as a conference paper at ICLR 2027}{Preprint}{}{\errmessage{Preprint header patch failed}}
\patchcmd{\@maketitle}{\LARGE\sc}{\Large\sc}{}{\errmessage{Title size patch failed}}
\patchcmd{\@maketitle}{\vskip 0.3in minus 0.1in}{\vskip 0.1in}{}{\errmessage{Title spacing patch failed}}
\makeatother
\begin{document}
\raggedbottom
\vspace*{-8pt}
\maketitle

\begin{abstract}
Text-to-image models have made remarkable progress in visual fidelity and instruction following, yet complex compositional generation remains challenging. Existing agentic approaches address this limitation through iterative generation and editing, adapting their actions based on interaction history.
However, they typically fix the continuation state (\emph{i.e.}, the image from which refinement proceeds) to the latest image.
Since iterative editing is not necessarily monotonic, the agent may continue refining from the latest image even when it has already degraded. 
In this work, we introduce \textbf{GenRevisit}, an agentic framework that extends action selection to \emph{joint continuation-state and action selection}. 
The agent can not only use interaction history to adjust editing instructions but also select an image candidate to edit, jointly deciding \emph{where to continue from and how to refine}.
We train GenRevisit with SFT followed by agentic reinforcement learning. To address the coarse credit assignment of final outcome reward in GRPO, which assigns the same trajectory-level advantage to all actions, we further propose Normalized Action Credit Assignment (NACA) to redistribute positive advantage according to each action's contribution to reducing the remaining objective. On GenEval2, GenRevisit achieves GM scores of 84.3 and 74.4 with Qwen-Image-2512 and Z-Image-Turbo, surpassing the strongest baselines using the same generators by 12.7 and 10.9 points respectively. GenRevisit achieves state-of-the-art performance on GenEval, GenEval++, and GenEval2 among baselines. 
\end{abstract}

\section{Introduction}
\label{sec:intro}
Recent text-to-image (T2I) models have made substantial progress in visual fidelity and instruction following~\citep{rombach2022highresolution,saharia2022photorealistic,wu2025qwenimage}, while satisfying object counts, attributes, and spatial relations in a single generation attempt remains challenging~\citep{ghosh2023geneval,kamath2025geneval2}.  To address this challenge, Iterative generation uses feedback on intermediate results to guide subsequent generation and editing~\citep{wan2025maestro,jaiswal2026iterative,li2025reflectdit}. As iterative generation produces multiple intermediate outcomes, the appropriate corrective action may vary from round to round. Fixed refinement procedures may therefore struggle to adapt to the evolving generation process, motivating an agentic sequential decision-making formulation in which the agent learns to select the next operation based on the accumulated interaction history~\citep{liu2026navigator}.

\begin{figure}[!t]
    \centering
    \includegraphics[width=0.95\linewidth]{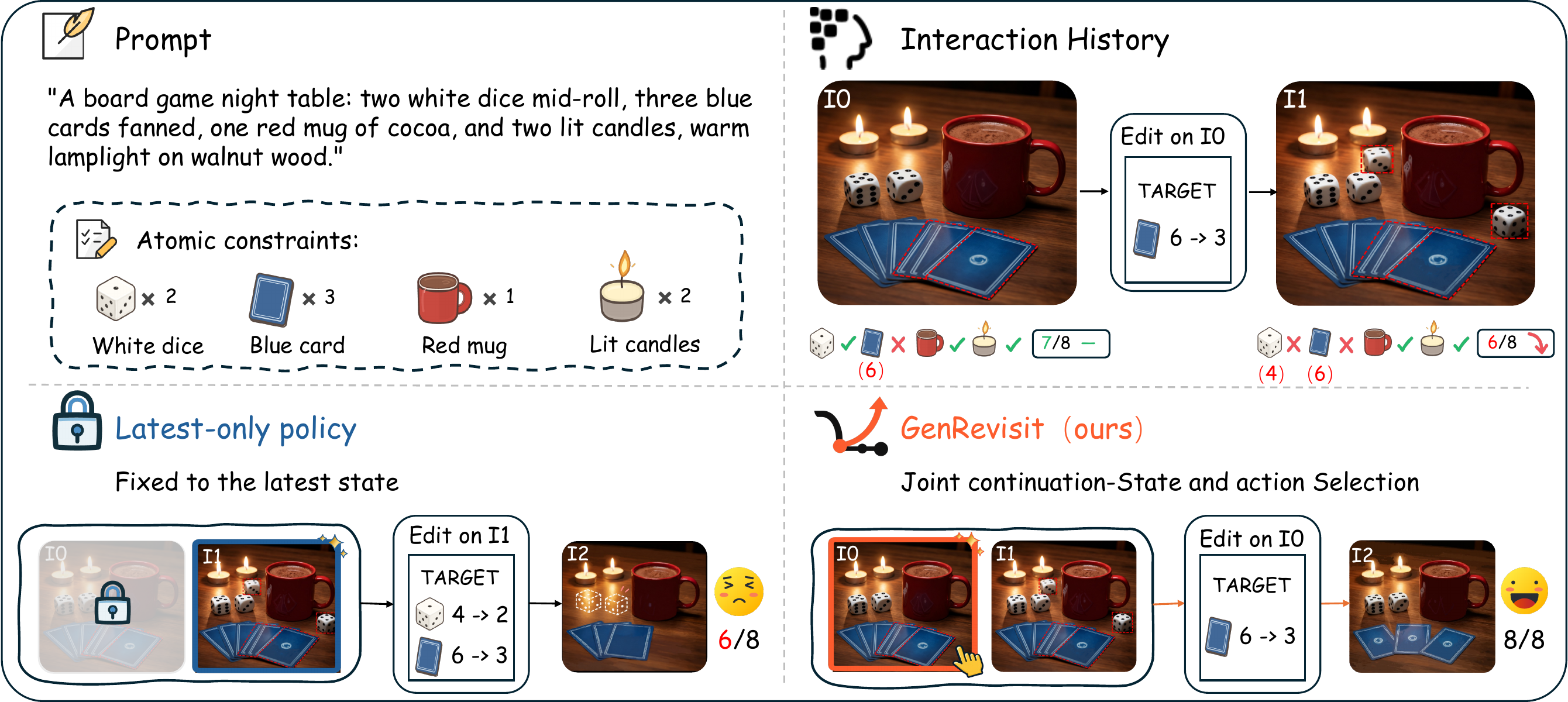}
    \caption{An illustrative comparison between latest-state refinement
    and our GenRevisit.}
    \label{fig:teaser}
\end{figure}

However, these policies mainly focus on learning what action to take next. An editing operation requires both a source image (\emph{i.e.}, the image from which refinement proceeds) and an editing instruction, but the source is often predetermined by the workflow. Although the interaction history records what action has attempted on each state and the resulting feedback, it works only for revising the instruction based on the fixed state if we just learn the action selection policy. This design has room for improvement because if we can't jointly decide continuation state and action to do, we may miss more promising state to continue for better results. 
For example, as illustrated in Figure~\ref{fig:teaser}, an edit to $I_0$ produces $I_1$ but violates previously satisfied constraints. A latest-only policy then gains additional feedback, yet its next edit must still start from the degraded image $I_1$. The agent may thus spend its limited budget restoring constraints that remain satisfied in $I_0$. This highlights the need for a policy that learns to jointly select a continuation state and specify an editing instruction tailored to it, rather than only adapting the instruction to a fixed source.

To this end, we introduce GenRevisit, a framework that expands decision space from action selection to joint continuation-state and action selection. We treat image candidates accumulated in the interaction history as candidate continuation states, each corresponding to a possible starting point for further refinement. 
At each decision step, GenRevisit uses a multimodal planner to choose the next action from generation, editing, querying, and submission, based on decision context.
When it edits, it jointly decides a source image and an editing instruction to correct remaining errors while preserving satisfied constraints. This allows the planner to revisit a stronger image candidate and open new refinement branches, rather than being restricted to the latest image.

We train our proposed GenRevisit 
in two stages. First, we collect multi-round trajectories with GPT-5.5 and initialize the planner by supervised fine-tuning (SFT).  We supervise actions that improve the image, including recovery from degraded states. Failed actions are excluded from the SFT loss but kept in the history with their feedback, because they still inform later decisions. We then use reinforcement learning (RL) with GRPO \cite{shao2024deepseekmath} to further improve the planner through its own interactions. The terminal reward measures how well the final submitted image satisfies the task constraints. However, the standard GRPO assigns the same advantage to every action in a trajectory, so a positive trajectory advantage can reinforce actions that worsen the image. To address this issue, we propose Normalized Action Credit Assignment (NACA). In positive-advantage trajectories, NACA redistributes credit among generation and editing actions based on the fraction of the remaining utility gap that each action closes.

Our experiments demonstrate the effectiveness of GenRevisit from three perspectives: \textbf{(i) Compositional generation:} GenRevisit achieves 84.3 GM on GenEval2 \cite{kamath2025geneval2}, exceeding the strongest agentic baseline by 12.7 points. %
Results on GenEval~\citep{ghosh2023geneval} and GenEval++~\citep{ye2025echo4o} further support cross-benchmark generalization. \textbf{(ii) Cross-generator transfer:} We use GenRevisit with Z-Image-Turbo \cite{zimage2025} without additional training. It also outperforms the strongest agentic baseline with the same generator by 10.9 points on GenEval2. \textbf{(iii) Benefit of joint selection:} Joint selection outperforms latest-only and best-only rules by 5.0 and 2.8 GM points, in matched-budget teacher continuations. The policy trained on joint demonstrations achieves 81.5 GM on GenEval2, exceeding its best-only-trained counterpart by 4.0 points.

Our contributions are summarized as follows:
\begin{itemize}
    \item We propose GenRevisit, a history-conditioned framework for joint continuation-state and action selection, making the choice of where to continue an explicit policy decision rather than a predetermined starting condition.
    \item We combine action-level SFT with RL and introduce NACA, which redistributes image-action credit in positive-advantage trajectories according to normalized local progress.
    \item GenRevisit achieves state-of-the-art performance on GenEval,  GenEval++ and GenEval2 among baselines, surpassing the  best GenEval2 results for Qwen-Image-2512 and Z-Image-Turbo by 12.7 and 10.9 points respectively, showing strong compositional generation.
\end{itemize}

\section{Related Work}
\label{sec:related-work}
\paragraph{Compositional evaluation.}
As text-to-image models are increasingly expected to satisfy constraints involving multiple objects, attributes, and relations simultaneously, compositional faithfulness has become a core test of generation capability. Earlier benchmarks such as GenEval, T2I-CompBench, and ConceptMix organize evaluation around predefined compositional dimensions~\citep{ghosh2023geneval,huang2023t2icompbench,wu2024conceptmix}, whereas TIFA, Davidsonian Scene Graph, and GenEval 2 move toward prompt-specific atomic verification by decomposing each request into individually verifiable rubrics~\citep{hu2023tifa,cho2024dsg,kamath2025geneval2}. Inspired by this paradigm, we use constraint-level atomic judgments as fine-grained environment feedback for multi-round agentic generation, allowing decisions to be grounded in how individual constraints evolve across attempts.

\paragraph{Iterative image generation.}
Iterative generation has progressed from prompt rewriting, repeated editing, and visual reflection~\citep{wan2025maestro,jaiswal2026iterative,li2025reflectdit} to closed-loop systems that evaluate intermediate results and invoke generation or editing tools accordingly~\citep{yang2026m3,he2026gems,jiang2026genagent}. More recent approaches make this feedback loop increasingly explicit: Qwen-Image-Agent incorporates verification feedback into subsequent refinement, while R3 organizes feedback-driven retry and repair~\citep{zhang2026qwenimageagent,wang2026r3}. Generation Navigator further formulates iterative generation as state-dependent action selection, learning when to stop, refine, or regenerate~\citep{liu2026navigator}. We build on this progression by extending the decision from what to do next to which source state to act from: historical candidates are retained as selectable branching points, allowing the policy to jointly choose a source state and the subsequent operation.

\section{Preliminaries}
\label{sec:preliminaries}

Agentic image generation can be viewed as a sequential decision process in which a planning policy interacts with external tools. Given a prompt specifying multiple compositional requirements, the objective is to produce an image that satisfies these requirements within a total budget of image generation and editing calls. 
At each decision step, the planner selects an action according to its
planning policy, conditioned on the current decision context.
The environment executes the action and updates the decision context
with the resulting observations and feedback.
We formalize this interaction in terms of states, actions, and trajectories.

\paragraph{States and actions.}
Let $x$ denote the input prompt and $B$ the total budget of image generation and editing calls. At decision step $t$, the interaction state $s_t$ denotes the decision context available to the planning policy, including $x$, the available history of actions, images, and feedback, and the remaining budget. The policy $\pi_\theta$, parameterized by $\theta$, selects an executable action as
\[
a_t \sim \pi_\theta(\cdot \mid s_t).
\tag{1}
\]
Each action specifies an operation and its associated parameters. Here, $s_t$ represents the overall decision context rather than an individual image candidate.
\paragraph{Interaction trajectories.}
The environment executes $a_t$ and incorporates the resulting observations and feedback into the next decision state $s_{t+1}$. The sequence of interactions forms a trajectory
\[
\tau=(s_0,a_0,s_1,a_1,\ldots,s_{T-1},a_{T-1},s_T),
\tag{2}
\]
where $T$ is the number of decision steps. The interaction terminates when a final image is selected as the output or another prescribed termination condition is met. The total number of generation and editing calls is bounded by $B$ and need not equal $T$.

While $s_t$ denotes the complete interaction state available for decision making, image candidates accumulated in the interaction history serve as candidate continuation states, each representing a possible starting point for further refinement. We refer to the candidate selected for this purpose as the \emph{continuation state}. Whether the continuation state is fixed to the latest candidate or explicitly selected from the interaction history is determined by the action interface. Section~\ref{sec:history-conditioned-interaction}  describes a history-conditioned interaction framework that enables joint continuation-state and action selection.
\section{Method}
\label{sec:method}
\begin{figure}[!t]
    \centering
    \includegraphics[width=\linewidth]{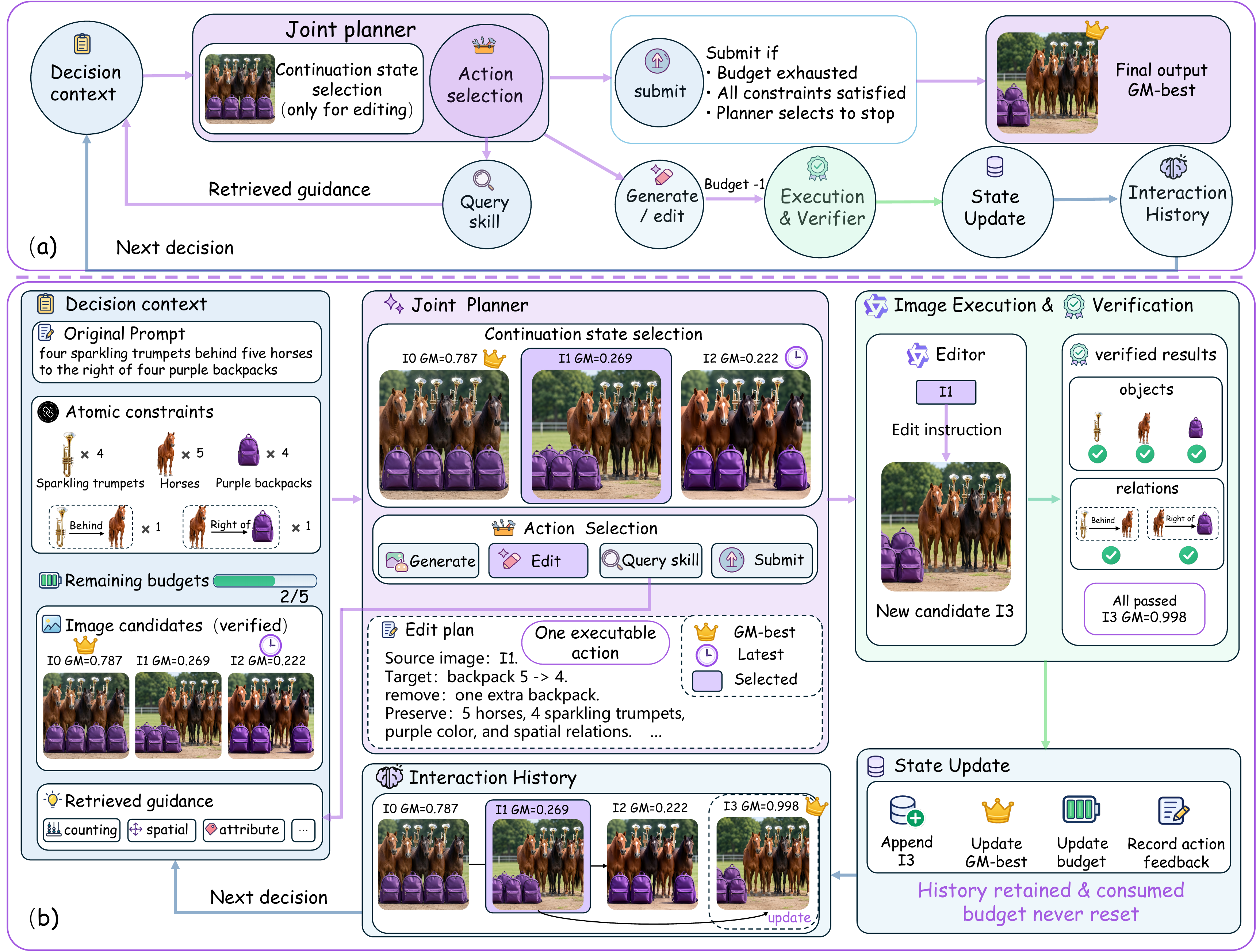}
    \caption{Overview of GenRevisit.
Panel (a) presents the overall workflow.
At each decision step, the multimodal planner selects the next
action based on the decision context and, when editing, jointly
selects a source image as the continuation state and a corresponding
edit specification.
The resulting image is verified and recorded in the interaction
history to inform subsequent decisions, while submission returns
the historical GM-best image.
Panel (b) illustrates an editing example.
The planner selects $I_1$, which is neither the latest candidate
$I_2$ nor the GM-best candidate $I_0$, and corrects the backpack
count while preserving the other constraints.
The resulting candidate $I_3$ passes all constraint-level checks
and becomes the new GM-best image.}
    \label{fig:pipeline}
\end{figure}

An overview of our GenRevisit is shown in Figure~\ref{fig:pipeline}. 
GenRevisit combines a history-conditioned interaction framework for
joint continuation-state and action selection with a two-stage planner
training procedure comprising supervised fine-tuning and reinforcement learning.
We detail the history-conditioned interaction framework in
Section~\ref{sec:history-conditioned-interaction}, supervised fine-tuning
in Section~\ref{sec:cold_start_sft}, and reinforcement learning with
normalized action credit assignment in Section~\ref{sec:rl_naca}.

\subsection{History-Conditioned Interaction}
\label{sec:history-conditioned-interaction}
\paragraph{Decision context.}
The decision context $s_t$ defined in Section~3 is instantiated
as shown in Figure~\ref{fig:pipeline}.
The compositional requirements of prompt $x$ are represented by
fixed atomic constraints $\mathcal{C}=\{c_j\}_{j=1}^{K}$,
each paired with a VQA-style verification question.
Historical image candidates are provided as visual inputs alongside
the actions that produced them and their verification feedback,
allowing the planner to assess potential editing sources.
Guidance returned by previous skill queries is also included
in the decision context.

\paragraph{Joint continuation-state and action selection.}
At each decision step, the agent produces one structured action
\[
a_t=(o_t,\eta_t)\sim\pi_\theta(\cdot\mid s_t),
\qquad
\eta_t=(k_t,u_t)\ \text{when }o_t=\textsc{Edit},
\]
where $o_t\in\mathcal{A}=\{\textsc{Generate},\textsc{Edit},\textsc{Retrieve},\textsc{Submit}\}$ denotes the operation type and $\eta_t$ contains its operation-specific arguments. Generation and editing specify the constraints to target and to preserve together with an executable instruction, and guidance retrieval requests constraint-specific advice. Submission returns the historical GM-best image and terminates the interaction.

For editing, $k_t$ indexes an available historical image candidate $I_{k_t}$
from the interaction history, which serves as the continuation state, and $u_t$ specifies the editing instruction
and the constraints to target and preserve.
This implements joint continuation-state and action selection: conditioned on \(s_t\), the planner jointly selects the source image \(I_{k_t}\) and its edit instruction \(u_t\) within a single structured action. The planner thus decides both where to continue refinement and how to address the selected image’s remaining errors while preserving satisfied constraints.

\paragraph{Execution and verification and update.}
The environment executes the selected action $a_t$, and for editing,
it applies the jointly selected edit specification $u_t$ to
the source image $I_{k_t}$.
Generation and editing each consume one image call.
Every resulting image, including the initial generation, is
evaluated against the fixed constraints $\mathcal{C}$, yielding
a GM score and constraint-level feedback.
After verification, the environment appends the new candidate,
its executed action, and feedback to the interaction history,
retaining the source image and editing instruction for edits.
It updates the historical GM-best candidate and remaining budget
to form the next decision context $s_{t+1}$.
Regressive attempts remain in the history, and editing an earlier
candidate creates a new branch without discarding subsequent
history or restoring consumed budget.

\subsection{Supervised Fine-Tuning For Cold Start}
\label{sec:cold_start_sft}
\paragraph{Teacher trajectorie construction.}

We use GPT-5.5 as a teacher agent to collect 1,000 multi-round trajectories in the history-conditioned interaction environment described in Section~\ref{sec:history-conditioned-interaction}, covering 1,000 prompts from the GenEval2 training set used by Flow-DPPO~\citep{ping2026flowdppo}. Training targets are selected from the teacher trajectories based on action outcomes. For edits, we retain actions that improve the image being edited. Degraded actions and their feedback remain in the context for later decisions while no imitation loss. The prompt-level split yields 3,473 training targets across 800 training prompts, with 100 prompts each reserved for validation and testing. Appendix~\ref{app:joint_sft} describes the full data construction process and reports the resulting dataset statistics.

\paragraph{Masked supervised fine-tuning.}
For each selected teacher action \(a_t\), we construct a training sequence \(z_j\) by serializing the corresponding interaction state \(s_t\) followed by the action response, with the associated images provided as conditioning inputs. The state \(s_t\) contains only information available before \(a_t\); the action’s outcome is used solely for offline target selection. For editing targets, the response includes both the image
choice \(k_t\) and the edit specification \(u_t\). We define a binary mask \(m_{j,\ell}\) that equals one on the action-response tokens and zero on all preceding context tokens, and minimize
\begin{equation}
\mathcal{L}_{\mathrm{SFT}}(\theta)
=
-\frac{1}{Z}
\sum_j \sum_{\ell}
m_{j,\ell}
\log \pi_\theta
\bigl(z_{j,\ell}\mid z_{j,<\ell}\bigr),
\qquad
Z=\sum_j\sum_{\ell}m_{j,\ell}.
\end{equation}
Here, \(\ell\) indexes token positions and \(Z\) is the total number of supervised tokens. Only the selected action response receives imitation loss, while prior actions and feedback, including failed attempts, remain available as conditioning context. The resulting SFT policy serves as the initialization for reinforcement learning in Section \ref{sec:rl_naca}.

\subsection{Reinforcement Learning with Normalized Action Credit}
\label{sec:rl_naca}

\paragraph{Terminal reward and trajectory advantage.}
For each prompt, we sample \(G\) trajectories under the policy \(\pi_{\mathrm{old}}\).
We define image utility as \(U(I)=P(I)+\alpha g(I)\), where
\(P(I)\) counts passed constraints and \(g(I)\in[0,1]\) is the
Soft-TIFA geometric mean of the verifier's correct-answer
probabilities across constraints~\citep{kamath2025geneval2}.
With \(\alpha=0.25\), the passed-constraint count takes priority,
while GM breaks ties. This priority is intentional: \(P(I)\) provides an explicit signal for
constraint completion, while \(g(I)\) retains sensitivity to continuous
improvements among images with the same pass count. For \(K\) constraints, the upper bound is
\(D=K+\alpha\).
For the image \(I_i^{\mathrm{sub}}\) submitted in trajectory \(i\),
the terminal reward and group-relative advantage are
\begin{equation}
R_i=U(I_i^{\mathrm{sub}}),
\qquad
A_i=\frac{R_i-\bar R}{\sigma_R},
\label{eq:naca_terminal_advantage}
\end{equation}
where \(\bar R\) and \(\sigma_R\) are the group mean and population
standard deviation. Failure handling and the exclusion of
zero-variance groups are described in Appendix~\ref{app:rl_rewards}.

\paragraph{Normalized action credit assignment.}
Standard GRPO assigns the same trajectory advantage to every action, potentially reinforcing regressive actions in positive-advantage trajectories. NACA redistributes positive advantage among generation and editing actions according to normalized local progress, without changing the terminal reward. For an image action \(a_{i,t}\) producing \(I_{i,t}\), the reference
utility \(U_{i,t}^{\mathrm{ref}}\) is the utility of the image being
edited, zero for initial generation, and the highest available
image utility before subsequent generation. We define
\begin{equation}
q_{i,t}
=
\frac{[U(I_{i,t})-U_{i,t}^{\mathrm{ref}}]_+}
     {D-U_{i,t}^{\mathrm{ref}}},
\label{eq:naca_normalized_progress}
\end{equation}
for \(U_{i,t}^{\mathrm{ref}}<D\), with \(q_{i,t}=0\) when
\(U_{i,t}^{\mathrm{ref}}=D\) and \([z]_+=\max(z,0)\).
This ratio measures the fraction of the remaining utility gap
closed by the action, rather than its absolute improvement.
An edit can therefore make positive progress without exceeding
the historical best GM score.

Let \(\mathcal I_i\) index valid, evaluated image actions and
\(L_{i,t}\) their trainable response-token counts.
For positive-advantage trajectories, effective positive-progress actions receive
\begin{equation}
w_{i,t}=\min(c_i,\lambda_i q_{i,t}),
\qquad
\sum_{t\in\mathcal I_i}L_{i,t}w_{i,t}
=\sum_{t\in\mathcal I_i}L_{i,t}.
\label{eq:naca_weights}
\end{equation}
Here, \(c_i\) caps the weights and \(\lambda_i\) scales them to preserve
the token-weighted total. Other image actions receive zero weight;
if no effective progress is available, all weights remain one.
Appendix~\ref{app:rl_allocation} specifies the effective-progress threshold,
cap, and numerical solver.
The action advantages are
\begin{equation}
\widetilde A_{i,t}
=
\begin{cases}
A_i w_{i,t}, & A_i>0\ \text{and}\ t\in\mathcal I_i,\\
A_i, & \text{otherwise}.
\end{cases}
\label{eq:naca_action_advantage}
\end{equation}
Nonpositive advantages remain unchanged, since weighting
negative advantages by progress could penalize improving
actions more strongly. Retrieval and submission also retain
\(A_i\).

\paragraph{Policy optimization.}
Each sampled action-response token inherits its action's
advantage, including the image choice and edit specification
for editing. At each RL round, the preceding checkpoint serves
as both \(\pi_{\mathrm{old}}\) and the frozen KL reference
\(\pi_{\mathrm{ref}}\).
Let \(\mathcal B\) contain the trajectories retained for an
optimizer step and \(\mathcal T_i\) their mask-selected response
tokens. For token \(y_{i,\ell}\) with multimodal prefix
\(H_{i,\ell}\), define
\(\rho_{i,\ell}=\pi_\theta(y_{i,\ell}\mid H_{i,\ell})/
\pi_{\mathrm{old}}(y_{i,\ell}\mid H_{i,\ell})\).
We maximize
\begin{equation}
\begin{aligned}
\mathcal J_{\mathrm{NACA}}(\theta)
={}&\frac{1}{|\mathcal B|}\sum_{i\in\mathcal B}
\sum_{\ell\in\mathcal T_i}
\Bigl[\min\bigl\{\rho_{i,\ell}\widetilde A_{i,\ell},\,
\operatorname{clip}(\rho_{i,\ell},1-\epsilon_-,1+\epsilon_+)
\widetilde A_{i,\ell}\bigr\}
-\beta\mathcal K_{i,\ell}\Bigr],
\end{aligned}
\label{eq:naca_policy_objective}
\end{equation}
where \(\epsilon_-\) and \(\epsilon_+\) are the clipping offsets,
and \(\beta\) weights the KL penalty \(\mathcal K_{i,\ell}\)
against \(\pi_{\mathrm{ref}}\).
The objective sums over response tokens and averages over
trajectories, without length normalization.
NACA changes advantages, not response masks: retained responses
with zero action advantage still contribute their KL terms.
Context and environment observations are inputs only.
Appendix~\ref{app:rl_details} gives the numerical solver,
KL estimator, and training details.

\section{Experiments}
\label{sec:experiments}
\subsection{Setup}
\label{sec:experimental-setup}
\paragraph{Implementation Details.}
We adopt Qwen3-VL-8B-Instruct~\citep{bai2025qwen3vl} as the planner backbone, with Qwen-Image-2512~\citep{qwen2025image2512} and Qwen-Image-Edit-2511~\citep{qwen2025imageedit2511} as frozen generation and editing tools. The skill library is predefined and remains unchanged during evaluation. The SFT training split contains 800 prompt groups and 3,473 selected action targets. SFT uses a learning rate of \(10^{-5}\), a global batch size of 32, and two epochs. The RL group size is four, with 1,280 sampled trajectories in total. Detailed dataset statistics, training settings, and tool configurations are provided in Appendices~\ref{app:training_settings}, \ref{app:joint_sft}, and~\ref{app:rl_details}.

\paragraph{Benchmark and Evaluation.}
We use GenEval2 as our primary benchmark for complex compositional generation and report its GM score. We additionally report overall scores on GenEval~\citep{ghosh2023geneval} and GenEval++~\citep{ye2025echo4o} to assess compositional generation beyond GenEval2. We also evaluate the same policy with Z-Image-Turbo on GenEval2 to test cross-generator transfer without additional training. We use greedy planner decoding with at most five generation or editing calls per prompt, including the initial generation, and evaluate the final submitted image. Baseline scores are taken from the cited sources under their reported configurations. Detailed evaluation protocols are provided in Appendix~\ref{app:evaluation}.

\subsection{Evaluations}
\begingroup
\hypersetup{hidelinks}
\definecolor{band}{HTML}{F3F3F3}
\definecolor{GenRevisit}{HTML}{EFEFEF}
\definecolor{gain}{HTML}{468B80}
\definecolor{loss}{HTML}{BB6963}
\definecolor{muted}{HTML}{555555}
\definecolor{rulegray}{HTML}{999999}
\newcommand{\metric}[3]{\makebox[1.9em][c]{#1}\ifnum#2=1\rlap{\textsuperscript{\normalfont\dag}}\fi\rlap{\if\relax\detokenize{#3}\relax\else\hspace{0.18em}{\fontsize{9.5}{11}\selectfont #3}\fi}}
\newcommand{\upgain}[1]{\textcolor{gain}{$\uparrow$#1}}
\newcommand{\downgain}[1]{\textcolor{loss}{$\downarrow$#1}}
\newcommand{\agent}[2]{\mbox{\raisebox{-2pt}[9pt][2pt]{\makebox[11pt][c]{#1}}\hspace{3.5pt}{\fontsize{11}{13}\selectfont #2}}}
\newcommand{\agentadjusted}[2]{\agent{#1}{#2}}
\newcommand{\claudeagent}[1]{\agent{\includegraphics[height=10pt,width=10pt,keepaspectratio]{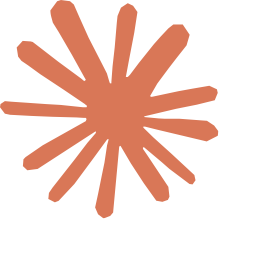}}{#1}}
\newcommand{\generator}[2]{\shortstack{#1\\[3.5pt]{\fontsize{11}{13}\selectfont #2}}}
\newcommand{\sectionrow}[1]{\rowcolor{band}\multicolumn{10}{@{}l@{}}{\hspace{5pt}\rule[-3.5pt]{0pt}{16pt}{\fontsize{12.5}{14}\selectfont\textcolor{muted}{\textsc{#1}}}}\\}
\setlength{\tabcolsep}{0pt}
\renewcommand{\arraystretch}{1.30}

\begin{table}[!t]
\centering
\caption{Main results on GenEval2. GM averages prompt-level geometric means of atom scores; Skill Avg. averages the five skill-wise AM scores. Bold and underlined scores mark the best and second-best results among (i) all direct-generation models and Qwen-Image-2512 workflows, or (ii) Z-Image-Turbo workflows alone. $\dagger$ marks the best external agentic baseline within each generator block; $\uparrow$/$\downarrow$ shows absolute score-point changes relative to it.}
\label{tab:geneval2-main}
\resizebox{\textwidth}{!}{%
\begin{minipage}{285mm}
\fontsize{13.5}{16}\selectfont
\begin{tabular}{@{}>{\centering\arraybackslash}p{32mm}@{\hspace{4mm}}>{\raggedright\arraybackslash}p{32mm}>{\hspace{3pt}\raggedright\arraybackslash\fontsize{12}{14}\selectfont}p{35mm}*{7}{>{\centering\arraybackslash}p{26mm}}@{}}
\toprule[0.8pt]
\multirow{2}{*}{\textbf{Generator}} & \multirow{2}{*}{\textbf{Agent}} & \multirow{2}{*}{\textbf{Method}} & \multicolumn{2}{c}{\textbf{Overall}$\uparrow$} & \multicolumn{5}{c}{\textbf{Skill-wise AM}$\uparrow$}\\
\cmidrule(lr){4-5}\cmidrule(lr){6-10}
 & & & \textbf{GM} & \textbf{Skill Avg.} & \textbf{Object} & \textbf{Attribute} & \textbf{Position} & \textbf{Count} & \textbf{Verb}\\
\midrule[0.45pt]
\sectionrow{Closed-source image generation models}
\multicolumn{2}{@{}l}{\hspace{5pt}Nano Banana~\citep{kamath2025geneval2}} & {\textcolor{muted}{Direct generation}} & \metric{44.6}{0}{} & \metric{\underline{83.5}}{0}{} & \metric{99.0}{0}{} & \metric{91.4}{0}{} & \metric{70.2}{0}{} & \metric{70.1}{0}{} & \metric{\textbf{86.7}}{0}{} \\
\multicolumn{2}{@{}l}{\hspace{5pt}Nano Banana 2~\citep{he2026gems}} & {\textcolor{muted}{Direct generation}} & \metric{63.4}{0}{} & \metric{81.7}{0}{} & \metric{97.0}{0}{} & \metric{\underline{95.1}}{0}{} & \metric{83.7}{0}{} & \metric{76.4}{0}{} & \metric{56.1}{0}{} \\
\addlinespace[2pt]
\sectionrow{Open-source image generation models}
\multicolumn{2}{@{}l}{\hspace{5pt}Bagel~\citep{deng2025bagel}} & {\textcolor{muted}{Direct generation}} & \metric{23.1}{0}{} & \metric{66.6}{0}{} & \metric{92.9}{0}{} & \metric{75.9}{0}{} & \metric{50.6}{0}{} & \metric{55.6}{0}{} & \metric{57.8}{0}{} \\
\multicolumn{2}{@{}l}{\hspace{5pt}Qwen-Image~\citep{wu2025qwenimage}} & {\textcolor{muted}{Direct generation}} & \metric{33.8}{0}{} & \metric{77.3}{0}{} & \metric{\underline{99.1}}{0}{} & \metric{85.6}{0}{} & \metric{60.2}{0}{} & \metric{70.3}{0}{} & \metric{\underline{71.1}}{0}{} \\
\multicolumn{2}{@{}l}{\hspace{5pt}Qwen-Image-2512~\citep{wu2025qwenimage}} & {\textcolor{muted}{Direct generation}} & \metric{29.0}{0}{} & \metric{62.0}{0}{} & \metric{97.9}{0}{} & \metric{65.2}{0}{} & \metric{58.1}{0}{} & \metric{66.9}{0}{} & \metric{22.1}{0}{} \\
\multicolumn{2}{@{}l}{\hspace{5pt}Z-Image-Turbo~\citep{he2026gems}} & {\textcolor{muted}{Direct generation}} & \metric{31.0}{0}{} & \metric{63.4}{0}{} & \metric{97.8}{0}{} & \metric{75.4}{0}{} & \metric{62.2}{0}{} & \metric{62.8}{0}{} & \metric{18.6}{0}{} \\
\addlinespace[4pt]
\sectionrow{Agentic generation workflows}
\addlinespace[3pt]
 & \tikzmark{genevaltwo-qstart} & Rewrite~\citep{he2026gems} & \metric{39.8}{0}{} & \metric{69.3}{0}{} & \metric{97.3}{0}{} & \metric{81.1}{0}{} & \metric{73.8}{0}{} & \metric{64.0}{0}{} & \metric{30.3}{0}{} \\
 &  & Promptist~\citep{hao2023promptist} & \metric{27.4}{0}{} & \metric{62.5}{0}{} & \metric{97.3}{0}{} & \metric{64.4}{0}{} & \metric{59.4}{0}{} & \metric{63.0}{0}{} & \metric{28.6}{0}{} \\
 & \multirow{-3}{*}{\agentadjusted{\includegraphics[height=10pt,width=10pt,keepaspectratio]{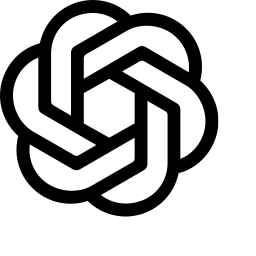}}{GPT-4}} & Search~\citep{ma2025inferencescaling} & \metric{42.1}{0}{} & \metric{67.5}{0}{} & \metric{98.3}{1}{} & \metric{71.1}{0}{} & \metric{66.1}{0}{} & \metric{73.9}{0}{} & \metric{28.2}{0}{} \\
\addlinespace[5pt]
 &  & Maestro~\citep{wan2025maestro} & \metric{48.0}{0}{} & \metric{72.4}{0}{} & \metric{97.3}{0}{} & \metric{84.9}{0}{} & \metric{80.9}{0}{} & \metric{68.0}{0}{} & \metric{31.0}{0}{} \\
 &  & Mind-Brush~\citep{he2026mindbrush} & \metric{32.1}{0}{} & \metric{--}{0}{} & \metric{--}{0}{} & \metric{--}{0}{} & \metric{--}{0}{} & \metric{--}{0}{} & \metric{--}{0}{} \\
 &  & SCOPE~\citep{ren2026scope} & \metric{46.3}{0}{} & \metric{--}{0}{} & \metric{--}{0}{} & \metric{--}{0}{} & \metric{--}{0}{} & \metric{--}{0}{} & \metric{--}{0}{} \\
 &  & CRAFT~\citep{kovalev2025craft} & \metric{66.3}{0}{} & \metric{82.4}{1}{} & \metric{97.9}{0}{} & \metric{91.3}{0}{} & \metric{89.9}{0}{} & \metric{79.0}{0}{} & \metric{53.7}{1}{} \\
 &  & GEMS~\citep{he2026gems} & \metric{70.4}{0}{} & \metric{81.3}{0}{} & \metric{96.9}{0}{} & \metric{91.4}{0}{} & \metric{91.4}{0}{} & \metric{\underline{81.3}}{1}{} & \metric{45.5}{0}{} \\
 & \multirow{-6}{*}{\agentadjusted{\includegraphics[height=10pt,width=10pt,keepaspectratio]{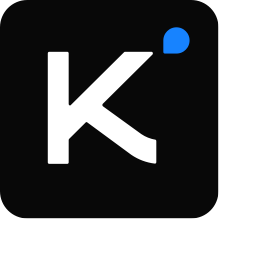}}{Kimi-K2.5}} & GenRouter~\citep{chen2026genrouter} & \metric{\underline{71.6}}{1}{} & \metric{81.8}{0}{} & \metric{98.3}{1}{} & \metric{92.3}{1}{} & \metric{\underline{91.6}}{1}{} & \metric{80.7}{0}{} & \metric{46.3}{0}{} \\
\addlinespace[5pt]
 &  & Base policy & \metric{49.7}{0}{} & \metric{74.6}{0}{} & \metric{97.9}{0}{} & \metric{88.9}{0}{} & \metric{64.4}{0}{} & \metric{72.4}{0}{} & \metric{49.6}{0}{} \\
\multirow{-11}{32mm}[5pt]{\centering\generator{\includegraphics[height=14pt,width=14pt,keepaspectratio]{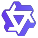}}{Qwen-Image-2512}} & \tikzmark{genevaltwo-qend}\multirow{-2}{*}{\agent{\includegraphics[height=10pt,width=10pt,keepaspectratio]{assets/geneval2-agent/qwen.pdf}}{Qwen3-VL-8B}} & \cellcolor{GenRevisit}\textbf{GenRevisit} & \cellcolor{GenRevisit}\metric{\textbf{84.3}}{0}{\upgain{12.7}} & \cellcolor{GenRevisit}\metric{\textbf{86.7}}{0}{\upgain{4.3}} & \cellcolor{GenRevisit}\metric{\textbf{99.4}}{0}{\upgain{1.1}} & \cellcolor{GenRevisit}\metric{\textbf{97.2}}{0}{\upgain{4.9}} & \cellcolor{GenRevisit}\metric{\textbf{92.5}}{0}{\upgain{0.9}} & \cellcolor{GenRevisit}\metric{\textbf{90.3}}{0}{\upgain{9.0}} & \cellcolor{GenRevisit}\metric{53.9}{0}{\upgain{0.2}} \\
\addlinespace[5pt]
\arrayrulecolor{rulegray}\midrule[0.35pt]\arrayrulecolor{black}
\addlinespace[4pt]
 & \tikzmark{genevaltwo-zstart} & GEMS~\citep{he2026gems} & \metric{\underline{63.5}}{1}{} & \metric{79.2}{0}{} & \metric{97.9}{0}{} & \metric{92.4}{0}{} & \metric{85.4}{0}{} & \metric{79.0}{0}{} & \metric{41.1}{0}{} \\
 & \multirow{-2}{*}{\agentadjusted{\includegraphics[height=10pt,width=10pt,keepaspectratio]{assets/geneval2-agent/kimi.png}}{Kimi-K2.5}} & GenRouter~\citep{chen2026genrouter} & \metric{57.6}{0}{} & \metric{76.4}{0}{} & \metric{97.9}{0}{} & \metric{89.5}{0}{} & \metric{76.8}{0}{} & \metric{80.3}{0}{} & \metric{37.4}{0}{} \\
\addlinespace[5pt]
 & \claudeagent{Sonnet 4.5} & \mbox{\fontsize{11}{13}\selectfont COMFYCLAW~\citep{li2026comfyclaw}} & \metric{62.0}{0}{} & \metric{70.3}{0}{} &  \metric{96.2}{0}{} & \metric{85.4}{0}{} & \metric{72.2}{0}{} & \metric{66.8}{0}{} & \metric{32.2}{0}{}  \\
\addlinespace[5pt]
\multirow{-4}{32mm}[3pt]{\centering\generator{\includegraphics[height=14pt,width=14pt,keepaspectratio]{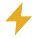}}{Z-Image-Turbo}} & \tikzmark{genevaltwo-zend}\agent{\includegraphics[height=10pt,width=10pt,keepaspectratio]{assets/geneval2-agent/qwen.pdf}}{Qwen3-VL-8B} & \cellcolor{GenRevisit}\textbf{GenRevisit} & \cellcolor{GenRevisit}\metric{\textbf{74.4}}{0}{\upgain{10.9}} & \cellcolor{GenRevisit}\metric{79.4}{0}{} & \cellcolor{GenRevisit}\metric{98.1}{0}{} & \cellcolor{GenRevisit}\metric{96.6}{0}{} & \cellcolor{GenRevisit}\metric{85.9}{0}{} & \cellcolor{GenRevisit}\metric{74.0}{0}{} & \cellcolor{GenRevisit}\metric{42.6}{0}{} \\
\bottomrule[0.8pt]
\end{tabular}%
\begin{tikzpicture}[remember picture,overlay]
\draw[draw=rulegray,line width=0.35pt] ([xshift=-2mm,yshift=7.5pt]pic cs:genevaltwo-qstart) -- ([xshift=-2mm,yshift=-3pt]pic cs:genevaltwo-qend);
\draw[draw=rulegray,line width=0.35pt] ([xshift=-2mm,yshift=7.5pt]pic cs:genevaltwo-zstart) -- ([xshift=-2mm,yshift=-3pt]pic cs:genevaltwo-zend);
\end{tikzpicture}
\end{minipage}
}
\end{table}
\arrayrulecolor{black}
\endgroup

\paragraph{Main results on GenEval2.} Table 1 shows that our method achieves 84.3 GM with Qwen-Image-2512, exceeding the strongest agentic baseline using the same generator by 12.7 points. It also attains the highest Skill Avg. of 86.7 and leads in four of the five skill categories. Counting shows the largest skill-wise gain, reaching 90.3 and exceeding the best competing score in this category by 9.0 points. Verb remains a comparatively weak category at 53.9. These results are obtained with an average of 2.62 image calls per prompt across all 800 prompts. With Z-Image-Turbo, the same policy achieves 74.4 GM without additional training, exceeding the strongest agentic baseline using this generator by 10.9 points. This result shows that the learned policy remains effective when paired with a different generator.

\paragraph{Cross-Benchmark Evaluation.} As shown in Table 2, our method achieves the highest overall scores on GenEval~\citep{ghosh2023geneval} and GenEval++~\citep{ye2025echo4o}, at 0.9844 and 0.8929, respectively. The strong counting performance on GenEval2 is also evident on GenEval (0.9875) and GenEval++ (1.000). These results suggest that the learned policy can guide compositional generation beyond GenEval2. Detailed per-category results and evaluation protocols are provided in Appendices~\ref{app:cross_benchmark_details} and~\ref{app:evaluation}.
\begingroup
\definecolor{crossband}{HTML}{F3F3F3}
\definecolor{crossours}{HTML}{EFEFEF}
\definecolor{crossmuted}{HTML}{555555}
\setlength{\tabcolsep}{0pt}
\renewcommand{\arraystretch}{1.30}
\newcommand{\crosssection}[1]{\rowcolor{crossband}\multicolumn{4}{@{}l@{}}{\hspace{4pt}\rule[-3.5pt]{0pt}{16pt}\textcolor{crossmuted}{\textsc{#1}}}\\}
\begin{table}[!t]\centering
\caption{Cross-benchmark evaluation on GenEval and GenEval++.}
\label{tab:cross-benchmark}\label{tab:geneval-ood}\label{tab:geneval-plus-plus}
\resizebox{\textwidth}{!}{%
\begin{minipage}{285mm}
\fontsize{12}{14}\selectfont
\begin{minipage}[t]{139mm}
\centering\textbf{(a) GenEval}\par\vspace{5pt}
\begin{tabular}{@{}>{\hspace{4pt}\raggedright\arraybackslash}p{39mm}>{\centering\arraybackslash}p{37mm}>{\centering\arraybackslash}p{46mm}>{\centering\arraybackslash}p{17mm}@{}}
\toprule[0.8pt]
\textbf{Method} & \textbf{Agent} & \textbf{Generator} & \textbf{Overall}$\uparrow$\\
\midrule[0.45pt]
\crosssection{Direct generation}
FLUX.1-dev~\citep{zhuo2025reflectionflow,jiang2026genagent} & -- & FLUX.1-dev & 67.0 \\
Qwen-Image~\citep{wu2025qwenimage,jiang2026genagent} & -- & Qwen-Image & 87.0 \\
Qwen-Image-2512$^{\dagger}$~\citep{he2026gems} & -- & Qwen-Image-2512 & 74.0 \\
Z-Image-Turbo$^{\dagger}$~\citep{he2026gems} & -- & Z-Image-Turbo & 82.0 \\
\addlinespace[3pt]\crosssection{Agentic generation and refinement}
GEMS~\citep{he2026gems} & Kimi K2.5 & Qwen-Image-2512 & 79.0 \\
GEMS~\citep{he2026gems} & Kimi K2.5 & Z-Image-Turbo & 86.0 \\
ReflectionFlow$^{\ddagger}$~\citep{zhuo2025reflectionflow} & Qwen2.5-VL-7B & FLUX.1-dev & 91.0 \\
RAISE~\citep{jiang2026raise} & Mistral-Small-3.2-24B & FLUX.1-dev + Kontext & 94.0 \\
CoIG~\citep{kim2025coig} & Gemini 2.5 Flash & Nano Banana & 94.8 \\
OctoT2I~\citep{jiang2026octot2i} & Qwen2-0.5B & 5-model pool & 96.0 \\
\addlinespace[3pt]\rowcolor{crossours}
\textbf{GenRevisit} & \textbf{Qwen3-VL-8B} & \textbf{Qwen-Image-2512} & \textbf{98.4} \\
\bottomrule[0.8pt]\end{tabular}\end{minipage}
\hfill%
\begin{minipage}[t]{139mm}
\centering\textbf{(b) GenEval++}\par\vspace{5pt}
\begin{tabular}{@{}>{\hspace{4pt}\raggedright\arraybackslash}p{39mm}>{\centering\arraybackslash}p{37mm}>{\centering\arraybackslash}p{46mm}>{\centering\arraybackslash}p{17mm}@{}}
\toprule[0.8pt]
\textbf{Method} & \textbf{Agent} & \textbf{Generator} & \textbf{Overall}$\uparrow$\\
\midrule[0.45pt]
\crosssection{Direct generation}
FLUX.1-dev~\citep{zhuo2025reflectionflow,jiang2026genagent} & -- & FLUX.1-dev & 32.5 \\
Qwen-Image~\citep{wu2025qwenimage,jiang2026genagent} & -- & Qwen-Image & 66.8 \\
GPT-Image-1.5~\citep{ye2026genclaw} & -- & GPT-Image-1.5 & 75.0 \\
Gemini-3.1 Flash-Image~\citep{ye2026genclaw} & -- & Gemini-3.1 & 77.5 \\
\addlinespace[3pt]\crosssection{Agentic generation and refinement}
ReflectionFlow$^{\ddagger}$~\citep{zhuo2025reflectionflow} & Qwen2.5-VL-7B & FLUX.1-dev & 36.1 \\
FiRe-GRPO (3rd)~\citep{kim2026fimr} & Janus-Pro-7B & Janus-Pro-7B & 49.1 \\
GenAgent~\citep{jiang2026genagent} & Qwen2.5-VL-7B & FLUX.1-dev & 56.1 \\
GenAgent~\citep{jiang2026genagent} & Qwen2.5-VL-7B & Qwen-Image & 72.5 \\
Mind-Brush~\citep{he2026mindbrush} & GPT-5.1 & Qwen-Image + Edit-2512 & 78.2 \\
GenClaw~\citep{ye2026genclaw} & Claude Opus 4.6 & Gemini-3.1 Flash-Image & 87.9 \\
\addlinespace[3pt]\rowcolor{crossours}
\textbf{GenRevisit} & \textbf{Qwen3-VL-8B} & \textbf{Qwen-Image-2512} & \textbf{89.3} \\
\bottomrule[0.8pt]\end{tabular}\end{minipage}

\par\vspace{5pt}
\end{minipage}%
}
\end{table}\endgroup

\begin{figure}[!t]
    \centering
    \includegraphics[width=0.95\linewidth]{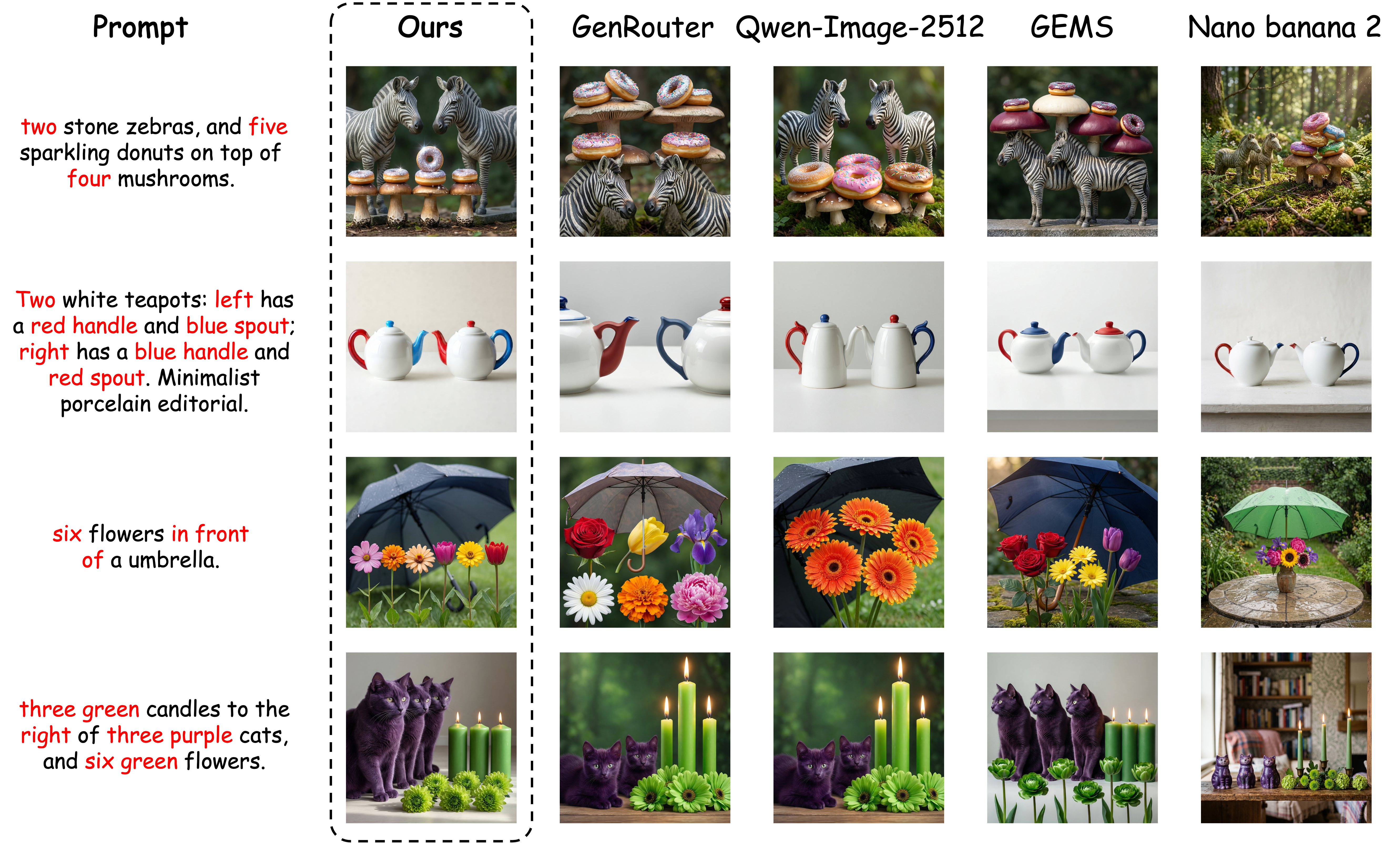}
    \caption{Qualitative comparison of GenRevisit and baseline models.}
    \label{fig:qualitative}
\end{figure}

\paragraph{Qualitative Comparison.}
As show in Figure~\ref{fig:qualitative}, GenRevisit presents qualitative comparisons on prompts involving counts, colors, attribute and relations. Our method more consistently satisfies these constraints, including the object counts, position-specific handle and spout colors of the teapots, and the counts and spatial arrangements of flowers, candles, and cats. These examples complement the quantitative results, illustrating improved compositional alignment while maintaining visual coherence. Additional outputs and an executed revisit trajectory are provided in Appendix~\ref{app:qualitative}.

\subsection{Ablation Study}
\label{sec:Ablation}
Table 3 examines the effects of our workflow and policy learning on GenEval2. Direct generation with Qwen-Image-2512 achieves 29.0 GM, while an untuned Qwen3-VL-8B planner reaches 49.7 using our workflow, with access to constraint-level feedback, interaction history, and historical images as editing sources. SFT raises GM to 81.5, with the largest skill-wise gains over the untuned planner in Position (+27.3) and Count (+16.4). This improvement highlights the value of learning the refinement policy rather than relying on an untuned planner to execute the workflow. Standard GRPO yields 80.3 GM, below the SFT policy’s 81.5. In contrast, GRPO with NACA reaches 84.3 GM, outperforming standard GRPO by 4.0 points and SFT by 2.8 points. The NACA variant also achieves the highest scores across all five skills among the evaluated variants. These results support redistributing positive trajectory credit according to the progress made by individual image actions.

\begingroup
\definecolor{ablationband}{HTML}{F3F3F3}
\definecolor{ablationfull}{HTML}{EFEFEF}
\definecolor{ablationgain}{HTML}{468B80}
\definecolor{ablationloss}{HTML}{BB6963}
\definecolor{ablationmuted}{HTML}{555555}
\definecolor{ablationrule}{HTML}{999999}

\setlength{\tabcolsep}{0pt}
\renewcommand{\arraystretch}{1.04}

\begin{table}[!t]
\centering
\caption{Ablation study on GenEval2.}
\label{tab:component-ablation}

\fontsize{7.5}{8.8}\selectfont

\begin{tabular}{
@{}
>{\hspace{3pt}\raggedright\arraybackslash}p{.31\linewidth}
*{2}{>{\centering\arraybackslash}p{.07\linewidth}}
>{\centering\arraybackslash}p{.10\linewidth}
*{5}{>{\centering\arraybackslash}p{.09\linewidth}}
@{}
}
\toprule[0.8pt]

\multirow{2}{*}{\textbf{Variant}}
& \multicolumn{3}{c}{\textbf{Overall}}
& \multicolumn{5}{c}{\textbf{Skill-wise AM}}
\\
\cmidrule(lr){2-4}
\cmidrule(lr){5-9}

& \textbf{GM}
& $\boldsymbol{\Delta}$\textbf{GM}
& \textbf{Skill Avg.}
& \textbf{Object}
& \textbf{Attribute}
& \textbf{Position}
& \textbf{Count}
& \textbf{Verb}
\\
\midrule[0.45pt]

\rowcolor{ablationband}
\mbox{Qwen-Image-2512}
& 29.0
& \textcolor{ablationmuted}{--}
& 62.0
& 97.9
& 65.2
& 58.1
& 66.9
& 22.1
\\

\mbox{Untuned Qwen3-VL-8B-Instruct$^{\dagger}$}
& 49.7
& \textcolor{ablationgain}{$+20.7$}
& 74.6
& 97.9
& 88.9
& 64.4
& 72.4
& 49.6
\\

\addlinespace[0.5pt]
\arrayrulecolor{ablationrule}
\midrule[0.35pt]
\arrayrulecolor{black}
\addlinespace[0.5pt]

\mbox{SFT only}
& 81.5
& \textcolor{ablationgain}{$+31.8$}
& 85.3
& 98.9
& 96.1
& 91.7
& 88.8
& 51.1
\\

\mbox{SFT + GRPO}
& 80.3
& \textcolor{ablationloss}{$-1.2$}
& 85.1
& 98.7
& 96.0
& 91.1
& 88.4
& 51.4
\\

\rowcolor{ablationfull}
\mbox{\textbf{SFT + NACA}}
& \textbf{84.3}
& \textcolor{ablationgain}{$+4.0$}
& \textbf{86.7}
& \textbf{99.4}
& \textbf{97.2}
& \textbf{92.5}
& \textbf{90.3}
& \textbf{53.9}
\\

\bottomrule[0.8pt]
\end{tabular}

\vspace{1pt}

\begin{minipage}{\linewidth}
\fontsize{7}{8.2}\selectfont
$^{\dagger}$Untuned Qwen3-VL-8B-Instruct uses our agent workflow.
$\Delta$GM is relative to the preceding row (NACA vs.\ standard GRPO).
\end{minipage}

\end{table}

\arrayrulecolor{black}
\endgroup

\subsection{Analysis}
\label{sec:core obeservation}
\label{sec:analysis}
Although we intuitively might assume that continuing to edit from the latest
state would yield better results, by analyzing the agent's decision
trajectories, we find it implicitly selects the continuation state
in three patterns: \textbf{(i)} following an edit that breaks satisfied constraints without resolving others, the agent often returns to an earlier, higher-scoring candidate; \textbf{(ii)} conversely the agent may continue from the latest image despite a lower overall score if it satisfies previously unmet constraints; \textbf{ (iii)} an earlier candidate that preserves constraints violated by later edits may be selected even if it is neither latest nor highest-scoring, providing a cleaner starting point for correcting the remaining errors. We further quantify the benefits of dynamic source selection through two experiments:

\textbf{Controlled teacher continuations.} To quantify the benefit of joint selection, we select 200 prefixes from a pool of 800 teacher trajectories where the task remains unsolved after three image calls. Each prefix provides a shared starting history for Joint, Latest-only, and Best-only, with two image calls remaining. All three use the same teacher and interaction protocol, differing only in edit-source access. Latest-only uses the latest image, Best-only uses the current highest-GM image, and Joint selects the source together with its editing instruction. Joint achieves a mean final GM of 55.2, compared with 52.4 for Best-only and 50.2 for Latest-only. Its improvement over Best-only is 2.8 points (paired 95\% CI: [0.46, 5.15]). Joint also increases the proportion of prefixes producing a new candidate above the shared prefix best from 57.0\% with Best-only to 65.0\%. These results support the value of joint selection beyond fixing the editing source to the current highest-scoring image.

\textbf{Learning from joint demonstrations.} A second experiment examines whether joint demonstrations lead to better performance after SFT. An additional 800 Best-only teacher trajectories are collected using the same prompts and initial images as the Joint trajectories. The two demonstration sets are used to train separate student policies via SFT. Both policies are then evaluated on the official GenEval2 evaluation set of 800 prompts under the same evaluation protocol used for our main results. The student trained on Joint demonstrations achieves 81.5 GM, compared with 77.5 for its Best-only-trained counterpart, an improvement of 4.0 points (paired 95\% CI: [2.81, 5.18]). The advantage is thus observed not only in controlled teacher continuations but also in the benchmark performance of the trained policies.

\section{Conclusion}
We introduce GenRevisit, which learns to jointly select a continuation state and an action. SFT retains failed actions and outcomes as recovery context rather than supervision. NACA assigns action-level credit using normalized local progress in non-monotonic trajectories. Experiments demonstrate improved compositional generation across benchmarks and cross-generator transfer without additional planner training. Our findings suggest that an image’s value as a starting point depends on both its current constraint satisfaction and what a suitable edit can achieve, highlighting the importance of learning source selection and editing together.

\clearpage
\ifnum\value{page}<11
  \setcounter{page}{11}
\fi
\bibliography{iclr2027_conference}
\bibliographystyle{unsrtnat}

\appendix
\makeatletter
\setlength{\@fptop}{0pt}
\setlength{\@fpsep}{12pt}
\makeatother
\raggedbottom
\clearpage
\AppActivateFormatting
\numberwithin{equation}{section}
\numberwithin{table}{section}
\numberwithin{figure}{section}
\numberwithin{algorithm}{section}
\phantomsection
\pdfbookmark[0]{Appendix Overview}{appendix.overview}
\begin{center}
{\large\scshape Appendix Overview}\par

\end{center}
\noindent Implementation details, training data, the continuation-state argument, and recorded cases are organized below. Section E reports the continuation protocols and cross-benchmark category results.
\vspace{5pt}
\begingroup
\small
\setlength{\parskip}{0pt}
\newcommand{\AppNavSection}[2]{\par\vspace{5pt}\noindent\hyperref[#1]{\textbf{\ref*{#1}\quad #2}}\hfill\hyperref[#1]{\pageref*{#1}}\par}
\newcommand{\AppNavSub}[2]{\noindent\hspace*{1.3em}\hyperref[#1]{\ref*{#1}\quad #2}\hfill\hyperref[#1]{\pageref*{#1}}\par}
\AppNavSection{app:interaction_protocol}{Experimental Setup and Interaction Protocol}
\AppNavSub{app:models}{Models and Frozen Components}
\AppNavSub{app:decision_inputs}{Decision Inputs, Actions, and Candidate Access}
\AppNavSub{app:training_settings}{Training Settings}
\AppNavSub{app:evaluation}{Benchmark and Evaluation Protocols}
\AppNavSection{app:joint_sft}{Teacher Data and Selective Supervision}
\AppNavSub{app:teacher_collection}{Prompt Selection and Teacher Collection}
\AppNavSub{app:sft_selection}{Action-Level Target Selection}
\AppNavSub{app:sft_masks}{Decision-Time Inputs and Loss Masks}
\AppNavSub{app:sft_statistics}{Data Splits and Corpus Statistics}
\AppNavSection{app:rl_details}{Reinforcement Learning Details}
\AppNavSub{app:rl_rounds}{Rollout Initialization and Policy References}
\AppNavSub{app:rl_rewards}{Terminal Rewards and Group Advantages}
\AppNavSub{app:rl_progress}{Reference Utility and Local Progress}
\AppNavSub{app:rl_allocation}{Token-Weighted Credit Allocation}
\AppNavSub{app:rl_optimization}{Response Masks and Policy Optimization}
\AppNavSection{app:historical_theory}{Optimization Potential of Historical Continuation}
\AppNavSub{app:joint_state_action}{Interaction State and Continuation State}
\AppNavSub{app:source_monotonicity}{Feasible-Set Monotonicity}
\AppNavSub{app:theory_scope}{Interpretation and Scope}
\AppNavSection{app:analysis}{Analysis Protocols and Additional Results}
\AppNavSub{app:teacher_continuations}{Controlled Teacher Continuations}
\AppNavSub{app:joint_demonstrations}{Learning from Joint Demonstrations}
\AppNavSub{app:student_continuations}{Historical-Source Access after Regression}
\AppNavSub{app:cross_benchmark_details}{Cross-Benchmark Category Results}
\AppNavSection{app:cases}{Case Studies and Qualitative Comparisons}
\AppNavSub{app:joint_best_case}{Joint and Best-only Continuations}
\AppNavSub{app:historical_latest_case}{Historical Continuation and Latest-only Refinement}
\AppNavSub{app:quality_potential_case}{Current Quality and Continuation Potential}
\AppNavSub{app:credit_case}{Action Credit on a Recorded Trajectory}
\AppNavSub{app:qualitative}{Additional Qualitative Outputs}
\AppNavSection{app:action_interfaces}{Prompts and Tool Interfaces}
\AppNavSub{app:prompt_contracts}{Teacher and Student Prompt Contracts}
\AppNavSub{app:action_schema}{Action Fields and Examples}
\AppNavSub{app:guidance_verifier}{Guidance Retrieval and Verifier Interfaces}
\AppNavSection{app:limitations}{Limitations}
\endgroup
\clearpage

\section{Experimental Setup and Interaction Protocol}
\label{app:interaction_protocol}
\AppBookmark{0}{app:interaction_protocol}{\thesection\space Experimental Setup and Interaction Protocol}
This appendix separates the planner's decisions from the information and rules supplied by the environment. We first describe the models, candidate access, and training settings. Appendix~\ref{app:joint_sft} specifies the teacher supervision, and Appendix~\ref{app:rl_details} gives the RL objective and credit-allocation algorithm. The training chain is Joint selective SFT followed by five rounds of NACA RL.

\subsection{Models and Frozen Components}
\label{app:models}
\AppBookmark{1}{app:models}{\thesubsection\space Models and Frozen Components}
The planner is initialized from Qwen3-VL-8B-Instruct~\citep{bai2025qwen3vl}, and GPT-5.5 supplies teacher trajectories. Generation and editing use separate, frozen backends: Qwen-Image-2512~\citep{qwen2025image2512} creates a new image, while Qwen-Image-Edit-2511~\citep{qwen2025imageedit2511} modifies the selected candidate. The environment routes the action to the corresponding backend. The planner specifies the image content and, for editing, the image to modify; it does not select a model, provider, sampling configuration, or image seed.

\begin{table}[!htbp]
\centering\small
\caption{Model roles and trainable components. Backend routing is controlled by the environment, not predicted in an action.}
\label{tab:models}
\begin{tabularx}{\linewidth}{@{}>{\raggedright\arraybackslash}p{.21\linewidth}>{\raggedright\arraybackslash}p{.31\linewidth}X@{}}
\toprule
\rowcolor{appTableHead}
\textbf{Role} & \textbf{Model or component} & \textbf{Use}\\
\midrule
Planner & Qwen3-VL-8B-Instruct & Language parameters are updated; vision tower and multimodal projector are frozen.\\
Teacher & GPT-5.5 & Collects multi-round demonstrations with executed actions and verifier feedback.\\
Generator & Qwen-Image-2512 & Produces a new candidate without an editing input.\\
Editor & Qwen-Image-Edit-2511 & Edits the candidate named in the action.\\
Verifier & Qwen3-VL-8B-Instruct & Frozen VQA evaluator; provides atomic feedback and Soft-TIFA scores.\\
Guidance & Fixed skill library & Returns requested guidance; does not choose the next action or image.\\
\bottomrule
\end{tabularx}
\end{table}

The cross-generator evaluation replaces the generation backend with Z-Image-Turbo without additional planner training. The generation and editing roles should be distinguished: an interface for replacing the generator does not, by itself, specify a different editing model. The exact evaluation-time tool binding belongs to the corresponding run configuration.

\subsection{Decision Inputs, Actions, and Candidate Access}
\label{app:decision_inputs}
\AppBookmark{1}{app:decision_inputs}{\thesubsection\space Decision Inputs, Actions, and Candidate Access}
A task consists of an original prompt, a fixed set of atomic constraints, and an image-call budget of five. Constraint identifiers and evaluator questions remain fixed throughout the episode. After each image action, the verifier records a status for every constraint and the correct-answer probabilities used to calculate Soft-TIFA GM. Statuses distinguish \texttt{pass}, \texttt{fail}, and \texttt{uncertain}; uncertainty is not silently replaced by a confident failure or success.

The Joint input presents all completed candidates in completion order, together with their images, evaluations, actual parent identifiers, and executed actions. Image pixels are supplied as multimodal message content and explicitly bound to candidate and artifact identifiers. They are not an additional field generated by the planner. Table~\ref{tab:decision_inputs} summarizes the information available for a decision; the exact interface fields are given in Appendix~\ref{app:action_interfaces}.

\begin{table}[!htbp]
\centering\small
\caption{Decision-time information in the Joint input. All entries are environment-provided inputs.}
\label{tab:decision_inputs}
\begin{tabularx}{\linewidth}{@{}>{\raggedright\arraybackslash}p{.25\linewidth}X@{}}
\toprule
\rowcolor{appTableHead}
\textbf{Component} & \textbf{Information}\\
\midrule
Task & Original prompt, fixed atomic requirements and questions, maximum image attempts.\\
Candidate history & Every completed candidate, its bound image, parent identifier, GM, constraint results, and the action that produced it.\\
Control & Current rankings, visible and editable candidate identifiers, legal submissions, and remaining image budget.\\
Retrieved guidance & Actual returned skill content and its associated target constraints.\\
Query control & Available skills, completed and pending queries, and remaining query allowance.\\
Action error, when present & The last unresolved format or validation error, used to request a new complete action.\\
\bottomrule
\end{tabularx}
\end{table}

\paragraph{One structured action.}
Each response contains exactly one of generation, editing, guidance retrieval, and submission. An editing response jointly specifies the candidate identifier, target constraints, preservation constraints, and executable instruction. These are parts of one action, not independent image-selection and instruction-generation stages. Targets and preservation constraints must be disjoint and refer to the fixed task. Preservation fields specify the content to retain during editing.

\paragraph{Access and ranking.}
Let candidate $j$ have GM $g_j$, passed-constraint count $P_j$, and completion index $j$. The Joint control ranks candidates lexicographically by $(g_j,P_j,-j)$. It also retains the underlying reducer's pass-first identifier, ranked by $(P_j,g_j,-j)$, under an explicitly different name. The GM-best identifier controls submission. The planner can edit any completed candidate, whether or not it is latest, GM-best, or uniquely satisfies a constraint. There is no prescribed fraction of historical edits and no requirement to avoid the current best.

\begin{table}[!htbp]
\centering\small
\caption{Candidate permissions under the GM-control interface. Rankings and permissions are recomputed after each completed image. This table defines the interfaces, not additional trained variants.}
\label{tab:candidate_access}
\begin{tabularx}{\linewidth}{@{}l X X l@{}}
\toprule
\rowcolor{appTableHead}
\textbf{Interface} & \textbf{Visible candidates} & \textbf{Editable candidates} & \textbf{Submission}\\
\midrule
Joint & All completed images & All completed images & GM-best\\
Best-only & All completed images & Current GM-best & GM-best\\
Latest-only & All completed images & Current latest & GM-best\\
\bottomrule
\end{tabularx}
\end{table}

\begin{appbox}[appBoxTeal]{Three distinct uses of image scores}
These control rankings are distinct from the RL utility $U(I)=P(I)+0.25g(I)$ and from the source-relative GM criterion for SFT selection. The control determines which submission is legal; SFT selection determines which demonstrated actions are imitated; RL utility determines rewards and local progress. None of these roles is substituted for another.
\end{appbox}

\paragraph{Budget and termination.}
The first generation is included in the five-image budget. A new generation creates a root candidate; an edit creates a child of the declared input image. Neither restarting nor editing an earlier image resets the budget or removes later observations. Guidance retrieval and submission do not create image candidates. Infrastructure retries are handled by the execution layer rather than recorded as fictitious low-scoring candidates.

With no completed image, the agent cannot edit or submit. Once the image budget is exhausted, it must submit the GM-best with the budget reason. Otherwise, if that candidate passes all constraints, it must submit with the all-pass reason. If neither condition holds, it may continue or submit with \texttt{no\_productive\_action\_remaining}. Budget exhaustion takes precedence when the last image is also all-pass. The planner decides whether to stop early when neither mandatory stopping condition holds. Submission uses the fixed GM-best rule.

\paragraph{History and guidance.}
The environment rebuilds persistent state from validated actions, execution artifacts, and evaluation events. Failed or ineffective executed edits remain available as evidence. Raw rejected output does not become persistent free-form memory, although an unresolved action error can be returned for correction. Candidate records carry the image history and executed actions, rather than a separate free-form memory summary. Internal transition records may contain more detail than the student input.

At most two successful skill queries are permitted between completed image evaluations. Pending queries and already active skills cannot be queried again. Returned guidance supports the planner's instruction but does not choose its editing input, alter candidate permissions, or add image calls. The six documented skills and their interface are listed in Appendix~\ref{app:guidance_verifier}.

\subsection{Training Settings}
\label{app:training_settings}
\AppBookmark{1}{app:training_settings}{\thesubsection\space Training Settings}
Selective SFT establishes the structured joint decision behavior. The resulting checkpoint initializes five RL rounds. Only the planner's language parameters are optimized; image backends, verification, and the planner's visual components remain frozen. Tables~\ref{tab:sft_training_settings} and \ref{tab:rl_training_settings} separate the supervision/export settings from the RL optimizer settings.

\begin{table}[!htbp]
\centering\small
\caption{SFT settings and supervision interface, aligned with the main-paper configuration.}
\label{tab:sft_training_settings}
\begin{tabularx}{\linewidth}{@{}>{\raggedright\arraybackslash}p{.40\linewidth}X@{}}
\toprule
\rowcolor{appTableHead}
\textbf{Setting} & \textbf{Value}\\
\midrule
Planner initialization & Qwen3-VL-8B-Instruct\\
Prompt membership & 800 train / 100 validation / 100 test\\
Teacher prompt pool & 1,000 prompts across the three splits\\
Learning rate / global batch size / epochs & $10^{-5}$ / 32 / 2\\
Selected action targets & 3,473 train / 437 validation / 413 test\\
Targets per exported record & One canonical assistant action\\
Training target & Complete action response, including image choice and edit specification\\
Context loss & Masked; task, images, previous actions, and feedback remain inputs\\
Data format & LLaMA-Factory action-response records\\
Supervision scope & Action response only; no separate reasoning target\\
Frozen modules & Vision tower, multimodal projector, external backends and verifier\\
\bottomrule
\end{tabularx}
\end{table}

\begin{table}[!htbp]
\centering\small
\caption{RL settings for the five-round training chain. Hardware is specified only for the final round where it was confirmed.}
\label{tab:rl_training_settings}
\begin{tabularx}{\linewidth}{@{}>{\raggedright\arraybackslash}p{.42\linewidth}X@{}}
\toprule
\rowcolor{appTableHead}
\textbf{Setting} & \textbf{Value}\\
\midrule
\rowcolor{appTableGroup}
\multicolumn{2}{@{}l}{\textbf{Training protocol}}\\
Initialization & Joint selective SFT checkpoint\\
Rounds / prompt groups per round & 5 / 64\\
Rollouts per group & 4\\
Collected trajectories & 256 per round; 1,280 total\\
Optimizer steps & 8 per round; 40 total\\
Configured update batch & 8 prompt groups, or 32 trajectories, before filtering\\
Data reuse / image budget & 1 epoch per round / 5 calls per trajectory\\
\midrule
\rowcolor{appTableGroup}
\multicolumn{2}{@{}l}{\textbf{Sampling and references}}\\
Temperature / top-$p$ / top-$k$ & 0.7 / 0.95 / disabled ($-1$)\\
Rollout policy & Preceding checkpoint, fixed during collection\\
KL reference & Frozen parent checkpoint, refreshed each RL round\\
\midrule
\rowcolor{appTableGroup}
\multicolumn{2}{@{}l}{\textbf{Optimization and credit}}\\
Optimizer / weight decay & AdamW / 0\\
Initial learning rate & $10^{-6}$, with linear decay within each round\\
Clipping offsets / KL coefficient & $(0.20,0.28)$ / $0.02$\\
Entropy coefficient / gradient clip & 0 / 1\\
Loss reduction & Trajectory mean, sampled-response token sum\\
Microbatch and accumulation & One action segment at a time; accumulate the step's assigned segments\\
Utility coefficient / nominal cap & $\alpha=0.25$ / $8$; effective cap can be higher\\
Nonpositive advantages & Broadcast unchanged\\
Credit preparation & After rollout, using execution events and response masks\\
Final-round distributed training & FSDP with four ranks\\
\bottomrule
\end{tabularx}
\end{table}

The 320 prompt-group instances need not correspond to 320 distinct prompts. The configured batch size is not a fixed gradient-accumulation count: the number of action segments depends on the sampled trajectories. Groups with effectively zero reward variance are excluded from optimizer data. The threshold, dynamic cap, and exact loss are defined in Appendix~\ref{app:rl_details}.

\subsection{Benchmark and Evaluation Protocols}
\label{app:evaluation}
\AppBookmark{1}{app:evaluation}{\thesubsection\space Benchmark and Evaluation Protocols}
GenEval2~\citep{kamath2025geneval2} is the primary compositional benchmark, with final image quality reported using Soft-TIFA GM. For the fixed $K$ atomic questions associated with an image $I$, let $p_k(I)$ be the verifier's probability of the correct answer. The documented score computation is
\begin{equation}
g(I)=\exp\!\left(\frac{1}{K}\sum_{k=1}^{K}\log\max\{p_k(I),10^{-300}\}\right).
\label{eq:app_gm_score}
\end{equation}
Missing, non-finite, or out-of-range probabilities are rejected before aggregation. GM is not the proportion of passed constraints. Discrete statuses support planning and pass-count computation, while the continuous probabilities define the GM score.

The reported GenEval2 datasets contain 800 prompts. The planner's ordinary evaluation decoding is greedy, with a maximum of five generation or editing calls including the initial generation. Evaluation scores refer to the selected submitted image rather than automatically to the last generated image. Cross-generator evaluation retains the planner weights and changes the generation backend without additional training.

The Qwen-backed evaluation reports the final submitted images for all 800 prompts, yielding 84.3 GM. We also rescore the same 800 submitted images with the official evaluator. On the GM$\times100$ scale, the absolute difference between the online and official mean scores is $0.00000082$, and the maximum per-image absolute difference is $0.00000437$. Thus, online scoring and official rescoring agree numerically on this evaluation set.

Execution validates action structure and source eligibility before dispatching image calls. Error handling uses action-validity feedback, deterministic target--preserve overlap resolution, and source-ID correction under the applicable source-eligibility rule. Editing instructions remain planner-generated, and continuation retains the executed prefix and remaining image budget. The current Joint prompt and the archived historical-access protocol are specified separately in Appendices~\ref{app:verbatim_prompts} and~\ref{app:student_continuations}.

GenEval~\citep{ghosh2023geneval} and GenEval++~\citep{ye2025echo4o} provide the cross-benchmark evaluations in the main paper. Their benchmark-specific results are reported separately from GenEval2 Soft-TIFA GM. External baseline values are taken from the cited publications, author-provided results, or reproductions under the configurations documented by those sources.

\clearpage
\section{Teacher Data and Selective Supervision}
\label{app:joint_sft}
\AppBookmark{0}{app:joint_sft}{\thesection\space Teacher Data and Selective Supervision}
The teacher data provide examples of complete decisions under the history-conditioned interface. Supervision is selected after actions have been executed, but each training input contains only the information available before its target action. This separates useful evidence from behavior that the student should imitate.

\subsection{Prompt Selection and Teacher Collection}
\label{app:teacher_collection}
\AppBookmark{1}{app:teacher_collection}{\thesubsection\space Prompt Selection and Teacher Collection}
GPT-5.5 collects multi-round trajectories on prompts from the GenEval2 training pool used by Flow-DPPO~\citep{ping2026flowdppo}. The original prompt split contains 800 training, 100 validation, and 100 test memberships. These are prompt counts, not counts of executed trajectories or selected action responses. The teacher prompt pool contains 1,000 prompts across these splits; the selected action responses are summarized in Table~\ref{tab:sft_statistics}.

Teacher collection uses the same four operation types and fixed generation/editing roles described in Appendix~\ref{app:interaction_protocol}. At an editing step, the teacher selects the image and the corresponding instruction together. It can use completed historical candidates, their evaluations, and earlier executed instructions, rather than restricting editing to a preselected best image.

Teacher collection and student training use different input wrappers. The teacher receives task, planner-context, candidate-observation, and continuation-control objects. The student receives a dedicated structured decision-time input. Both expose the GM-control permissions, but the teacher's longer instructions are not claimed to be the student's verbatim system prompt. In particular, the teacher's four-part editing guidance is a semantic structure inside one instruction string, not four additional action fields.

\subsection{Action-Level Target Selection}
\label{app:sft_selection}
\AppBookmark{1}{app:sft_selection}{\thesubsection\space Action-Level Target Selection}
Selective supervision retains individual actions using their observed outcomes. Let $g_{\mathrm{child}}$ denote the new image's GM, $g_{\mathrm{edit}}$ the GM of the image being edited, and $g_{\mathrm{archive}}$ the largest GM available before the action. The improvement tolerance is $\varepsilon_{\mathrm{SFT}}=10^{-9}$.

\begin{table}[!htbp]
\centering\small
\caption{Selection rules for positive SFT targets. All comparisons use the record before the candidate action and its actual execution outcome.}
\label{tab:sft_selection}
\begin{tabularx}{\linewidth}{@{}>{\raggedright\arraybackslash}p{.23\linewidth}X@{}}
\toprule
\rowcolor{appTableHead}
\textbf{Action} & \textbf{Selection rule}\\
\midrule
Initial generation & Retain the required first generation.\\
Editing & Retain when $g_{\mathrm{child}}>g_{\mathrm{edit}}+\varepsilon_{\mathrm{SFT}}$.\\
Subsequent generation & Retain when $g_{\mathrm{child}}>g_{\mathrm{archive}}+\varepsilon_{\mathrm{SFT}}$.\\
Submission & Retain the actually executed submission when it selects the current GM-best and uses a valid reason.\\
Guidance retrieval & Require a real returned response. The first subsequent executed image action must itself be selected, and its target constraints must intersect the query's target constraints.\\
\bottomrule
\end{tabularx}
\end{table}

An edit can be selected even when its output remains below the archive best, provided that it improves its own editing input. Constraint-level regressions are recorded separately and do not automatically veto a GM-positive target. This SFT rule is not the pass-count-first utility used in RL.

Actions that do not meet the selection rule are not imitation targets. Their validated execution records, images, and feedback remain available to later decisions. A GM-decreasing bridge action does not become a positive target merely because a later descendant succeeds. The student can therefore learn a later recovery without being trained to reproduce the earlier degrading action.

The export distinguishes a narrower recovery-positive label: an edit uses a non-latest image and exceeds the pre-action GM archive. Other retained actions are other positive. These labels should not be confused with editing an earlier candidate alone. Source-relative GM change, archive-relative GM change, and changes in passed constraints are recorded for audit; they are not additional planner outputs.

\subsection{Decision-Time Inputs and Loss Masks}
\label{app:sft_masks}
\AppBookmark{1}{app:sft_masks}{\thesubsection\space Decision-Time Inputs and Loss Masks}
Each selected target is one canonical assistant action paired with its decision-time input. The input is reconstructed from the preceding immutable event prefix and contains the associated candidate images in the documented order. The target's execution outcome is used only for offline selection; it is not inserted into that action's input. Previously completed unsuccessful actions remain legitimate context for later targets.

An exported row contains a system message, a user message with the rendered multimodal state, and one assistant action. Only the assistant response is supervised. For editing, this includes the candidate choice, target and preservation sets, and complete instruction. The export uses the action-response template with prompt training disabled. The number of image placeholders must equal the number of bound images; the binding preserves their identities rather than relying on an unlabelled positional guess.

For serialized sequence $z_j$, define $m_{j,\ell}=1$ on its selected response tokens and zero elsewhere. The objective is
\begin{equation}
\mathcal L_{\mathrm{SFT}}(\theta)
=-\frac{1}{Z}\sum_j\sum_\ell m_{j,\ell}
\log\pi_\theta(z_{j,\ell}\mid z_{j,<\ell}),
\qquad Z=\sum_j\sum_\ell m_{j,\ell}.
\label{eq:app_sft_loss}
\end{equation}
The associated images are conditioning inputs. Excluding a failed action from the target does not remove its earlier occurrence from a later input, and does not assign it a negative imitation loss. The resulting policy initializes the five-round RL training chain.

\subsection{Data Splits and Corpus Statistics}
\label{app:sft_statistics}
\AppBookmark{1}{app:sft_statistics}{\thesubsection\space Data Splits and Corpus Statistics}
All trajectories for the same prompt stay in the same split. Records are deduplicated using the split, rendered input, the identities of the ordered images, and canonical target action. Copying an identical shared prefix therefore does not create duplicate supervision weight; different real actions at the same state can remain distinct targets. Training samples action rows without an additional episode-level weighting rule.

\begin{table}[!htbp]
\centering\small
\caption{Selected targets in the current Joint SFT export. Counts are action responses, not distinct prompts or raw trajectories.}
\label{tab:sft_statistics}
\begin{tabularx}{\linewidth}{@{}Xrrrrr@{}}
\toprule
\rowcolor{appTableHead}
\textbf{Split} & \textbf{Generate} & \textbf{Edit} & \textbf{Retrieve} & \textbf{Submit} & \textbf{Total}\\
\midrule
Train & 961 & 1,076 & 584 & 852 & 3,473\\
Validation & 108 & 140 & 103 & 86 & 437\\
Test & 110 & 114 & 99 & 90 & 413\\
\midrule
Total & 1,179 & 1,330 & 786 & 1,028 & 4,323\\
\bottomrule
\end{tabularx}
\end{table}

The export records 6,858 image bindings and 3,418 unique image artifacts across the selected records. These quantities differ because the same image can appear in more than one decision context. Training uses 800 distinct prompts; action-response counts in Table~\ref{tab:sft_statistics} are not prompt counts. Validation and test submission-target totals likewise do not establish their raw collection sizes.

Validation and test rows are rendered and selected under the same compatibility rules as training rows. Their next-token losses measure conditional action modeling. They are not new online image-generation evaluations and should not be reported as final benchmark image quality.

\clearpage
\section{Reinforcement Learning Details}
\label{app:rl_details}
\AppBookmark{0}{app:rl_details}{\thesection\space Reinforcement Learning Details}

Starting from the Joint SFT checkpoint, we train the planner for five RL rounds with NACA. This section specifies the rollout state, per-round policy references, rewards, token-weighted action credit, and optimization loss. Training settings are summarized in Table~\ref{tab:rl_training_settings}.

\subsection{Rollout Initialization and Policy References}
\label{app:rl_rounds}
\AppBookmark{1}{app:rl_rounds}{\thesubsection\space Rollout Initialization and Policy References}

\paragraph{A group shares a task, not a generated image.}
A GRPO group contains $G=4$ complete trajectories with the same
prompt, fixed atomic constraints, and initial budget of five image
calls. Each trajectory begins before image generation, with no
completed image candidates or image evaluations. It executes its
own first \texttt{generate\_image} call and then its own subsequent
actions. No shared first image is supplied to the four trajectories.

For the initial generation, the image seed is
\texttt{base\_seed} + \texttt{candidate\_index}.
Here, \texttt{candidate\_index} distinguishes rollout candidates
in the group, not image-attempt identifiers within a trajectory.
Thus first images may differ. This rule does not assert that every
later image call uses that same seed formula. The first generation
counts toward each trajectory's five-image budget and is included
in NACA's image-action credit calculation.

This differs from a matched-prefix continuation experiment, which
copies an already executed and evaluated prefix into comparison
arms. Such a prefix can contain a shared first image or several
shared images, leaving four or fewer image calls. That continuation
protocol does not describe the initialization of these RL groups.

\paragraph{Training-round references.}
Let $\pi^{(0)}$ be the Joint selective SFT policy and $\pi^{(r)}$ the
checkpoint after RL round $r$, for $r=1,\ldots,5$.
Round $r$ collects trajectories using $\pi^{(r-1)}$ and initializes
the trainable planner from the same checkpoint. A frozen copy of
$\pi^{(r-1)}$ provides the KL reference:
\begin{equation}
\pi_{\mathrm{old}}^{(r)}=\pi^{(r-1)},
\qquad
\pi_{\mathrm{ref}}^{(r)}=\operatorname{freeze}(\pi^{(r-1)}).
\label{eq:app_parent_policies}
\end{equation}
The two policies have identical starting weights but different roles:
$\pi_{\mathrm{old}}^{(r)}$ supplies rollout probabilities, while
$\pi_{\mathrm{ref}}^{(r)}$ anchors the KL penalty. Their stored
log-probabilities remain fixed as $\pi_\theta$ is updated within the
round. The next round loads the newly trained checkpoint rather
than retaining the original SFT policy as its reference.

Each round completes collection before optimization. Recorded
execution events and sampled-response masks are then used to compute
NACA advantages, which are stored alongside the rollout data and
consumed by the optimizer. Each round uses one epoch and eight
optimizer steps. In the final round, the learning rates are
$10^{-6},8.75\times10^{-7},\ldots,1.25\times10^{-7}$.
Only planner language parameters are updated; the vision tower,
multimodal projector, image backends, and verifier remain frozen.

\subsection{Terminal Rewards and Group Advantages}
\label{app:rl_rewards}
\AppBookmark{1}{app:rl_rewards}{\thesubsection\space Terminal Rewards and Group Advantages}

For an image $I$, utility is $U(I)=P(I)+0.25g(I)$, where $P(I)$
counts passed constraints and $g(I)$ is the Soft-TIFA geometric mean.
The terminal reward uses the image explicitly submitted by the
planner, not an automatically substituted historical best.
The RL utility is intentionally distinct from the GM used for
benchmark reporting and GM-based image selection. For two candidate
images, its difference can be written as
\begin{equation}
\Delta U = \Delta P + 0.25\,\Delta g .
\label{eq:utility_vs_gm}
\end{equation}
Because \(g(I)\in[0,1]\), an increase of at least one in passed-constraint
count satisfies
\[
\Delta P\ge 1
\quad\Longrightarrow\quad
\Delta U \ge 1-0.25 = 0.75 > 0.
\]
Thus, the RL utility intentionally prefers an image that passes more
constraints, even when its GM is lower. When the pass count is unchanged,
the utility preserves the GM ordering. SFT target selection and final
image selection continue to use their specified GM-based criteria.
A policy-invalid trajectory receives $R_i=-1$ and remains in its prompt
group when computing the reward mean and variance.
Infrastructure failures are handled through retry or recovery;
they are not assigned fabricated image scores or invalid-policy
rewards.

\begin{table}[!htbp]
\centering
\small
\caption{Termination and failure handling during RL data preparation.}
\label{tab:rl_failure_handling}
\setlength{\tabcolsep}{5pt}
\renewcommand{\arraystretch}{1.14}
\begin{tabularx}{\linewidth}{@{}p{0.29\linewidth}X@{}}
\toprule
\rowcolor{appTableHead}
\textbf{Case} & \textbf{Handling}\\
\midrule
Normal submission & Reward is the utility of the submitted image.\\
Policy-invalid trajectory & Reward is $-1$ and participates in group statistics.
Retained sampled responses are trained if their group is not filtered.\\
Image backend, transport, or verifier failure & Retry or recover; do not
substitute a score or treat the failure as a policy violation.\\
Image budget exhausted, followed by valid submission & Score the submitted
image normally; budget exhaustion alone adds no penalty.\\
No valid submission & Use the recorded termination class to distinguish
policy invalidity from incomplete execution or infrastructure failure;
do not invent a submission.\\
Zero-variance group & Remove the group from optimizer data, including KL.\\
\bottomrule
\end{tabularx}
\end{table}

For a group of $G=4$ trajectories, define
\begin{equation}
\begin{aligned}
\bar R&=\frac{1}{G}\sum_{i=1}^{G}R_i,
&\sigma_R&=\sqrt{\frac{1}{G}\sum_{i=1}^{G}(R_i-\bar R)^2},\\
A_i&=
\begin{cases}
(R_i-\bar R)/\sigma_R, & \sigma_R>10^{-12},\\
0, & \text{otherwise}.
\end{cases}
\end{aligned}
\label{eq:app_group_advantage}
\end{equation}
No additive epsilon is used in the denominator. Groups at or below
the standard-deviation threshold are excluded from optimizer assignment, so
they contribute neither a policy-gradient term nor a KL term.

Policy-invalid cases include parsing errors, invalid candidate
references, and action-rule violations. In the final round, 16 of
256 trajectories were policy-invalid (6.25\%); these trajectories
were retained rather than removing their groups solely for
containing a policy error.

\subsection{Reference Utility and Local Progress}
\label{app:rl_progress}
\AppBookmark{1}{app:rl_progress}{\thesubsection\space Reference Utility and Local Progress}

Let $\mathcal V_{i,t}$ contain the evaluated image candidates
available before action $a_{i,t}$, and let $I_{i,k_{i,t}}$ be the
image selected for an edit. The reference utility is
\begin{equation}
U_{i,t}^{\mathrm{ref}}=
\begin{cases}
0, & \text{initial generation},\\
U(I_{i,k_{i,t}}), & \text{editing},\\
\max_{I\in\mathcal V_{i,t}}U(I), & \text{subsequent generation}.
\end{cases}
\label{eq:app_reference_utility}
\end{equation}
An edit is measured against the image it actually modifies.
A new generation has no selected image to edit, so its progress is
measured against the highest utility already available. This reference
uses $U$, not the GM-first submission ranking.

For a task with $K$ constraints, let $D=K+0.25$. For an action
producing $I_{i,t}$, normalized local progress is
\begin{equation}
q_{i,t}=
\begin{cases}
\displaystyle\frac{[U(I_{i,t})-U_{i,t}^{\mathrm{ref}}]_+}
{D-U_{i,t}^{\mathrm{ref}}}, & U_{i,t}^{\mathrm{ref}}<D,\\[0.6em]
0, & U_{i,t}^{\mathrm{ref}}=D,
\end{cases}
\label{eq:app_local_progress}
\end{equation}
where $[z]_+=\max(z,0)$. The normalization measures the fraction
of the remaining utility gap closed by an action, allowing both
initial generation and later repair to receive credit without
comparing their absolute gains alone.

Positive progress depends on utility, not on every individual
constraint or GM increasing. For example, an increase in passed
constraints can raise utility despite a lower GM. Conversely,
a non-increasing utility gives $q_{i,t}=0$. An edit can receive
positive progress without surpassing the historical best.
These are local comparisons, not estimates of an action's causal
contribution to the final submitted image.

\subsection{Token-Weighted Credit Allocation}
\label{app:rl_allocation}
\AppBookmark{1}{app:rl_allocation}{\thesubsection\space Token-Weighted Credit Allocation}

An image action is eligible when its evaluation has completed,
it is not policy-invalid, and
contains at least one trainable response token.
Let $\mathcal I_i$ index these actions and $L_{i,t}>0$ be the
number of mask-selected tokens in action $a_{i,t}$.
With $\delta=10^{-6}$, define
\begin{equation}
\begin{aligned}
T_i&=\sum_{t\in\mathcal I_i}L_{i,t},
&E_i&=\{t\in\mathcal I_i:q_{i,t}>\delta\},\\
T_i^+&=\sum_{t\in E_i}L_{i,t},
&Z_i&=\sum_{t\in\mathcal I_i}L_{i,t}q_{i,t}.
\end{aligned}
\label{eq:app_allocation_sets}
\end{equation}
Only positive trajectory advantages are redistributed.
If $E_i$ is empty or $Z_i\leq\delta$, all eligible image actions
use $w_{i,t}=1$. Otherwise, the effective cap and weights are
\begin{equation}
\begin{aligned}
c_i&=\max\!\left(8,\frac{T_i}{T_i^+}\right),\\
w_{i,t}&=
\begin{cases}
\min(c_i,\lambda_iq_{i,t}), & t\in E_i,\\
0, & t\in\mathcal I_i\setminus E_i,
\end{cases}
\qquad
\sum_{t\in\mathcal I_i}L_{i,t}w_{i,t}=T_i.
\end{aligned}
\label{eq:app_dynamic_cap}
\end{equation}
Thus, $8$ is a nominal cap, not a global maximum.
The constraint requires $c_iT_i^+\geq T_i$: otherwise even
saturating every positive-progress action cannot redistribute
the full token-weighted total. The implementation raises an
infeasible nominal cap to $T_i/T_i^+$ and records that adjustment. There is no additional fixed upper bound.

The scale $\lambda_i$ is found with 100 bisection iterations.
The initial bracket is
$[0,\max(1,c_i/\min_{t\in E_i}q_{i,t})]$.
Credit conservation is checked to tolerance
$\max(10^{-6},T_i10^{-10})$.
Algorithm~\ref{alg:naca_allocation} summarizes the allocation.
There is no positive weight floor: in the normal allocation branch,
actions with $q_{i,t}\leq10^{-6}$ receive zero weight.

\begin{algorithm}[!htbp]
\caption{NACA weights for one trajectory}
\label{alg:naca_allocation}
\small
\begin{algorithmic}[1]
\Require $A_i$, eligible image actions $\mathcal I_i$, token counts
$L_{i,t}$, local progress $q_{i,t}$; $\delta=10^{-6}$
\State Set $w_{i,t}\gets1$ for all $t\in\mathcal I_i$.
\If{$A_i\leq0$}
    \State \Return $w_i$
\EndIf
\State Compute $T_i$, $E_i$, $T_i^+$, and $Z_i$ using
Eq.~\eqref{eq:app_allocation_sets}.
\If{$E_i=\varnothing$ or $Z_i\leq\delta$}
    \State \Return $w_i$ \Comment{Uniform fallback}
\EndIf
\State $c_i\gets\max(8,T_i/T_i^+)$;
record a cap adjustment if $8<T_i/T_i^+$.
\State $\lambda_{\mathrm{lo}}\gets0$;
$\lambda_{\mathrm{hi}}\gets\max(1,c_i/\min_{t\in E_i}q_{i,t})$.
\For{$n=1,\ldots,100$}
    \State $\lambda\gets(\lambda_{\mathrm{lo}}+\lambda_{\mathrm{hi}})/2$
    \State $S\gets\sum_{t\in E_i}L_{i,t}\min(c_i,\lambda q_{i,t})$
    \If{$S<T_i$}
        \State $\lambda_{\mathrm{lo}}\gets\lambda$
    \Else
        \State $\lambda_{\mathrm{hi}}\gets\lambda$
    \EndIf
\EndFor
\State Form $w_i$ from the resulting scale using Eq.~\eqref{eq:app_dynamic_cap}.
\State Check $|\sum_{t\in\mathcal I_i}L_{i,t}w_{i,t}-T_i|
\leq\max(10^{-6},T_i10^{-10})$.
\State \Return $w_i$
\end{algorithmic}
\end{algorithm}

The resulting action advantage is
\begin{equation}
\widetilde A_{i,t}=
\begin{cases}
A_iw_{i,t}, & A_i>0\ \text{and}\ t\in\mathcal I_i,\\
A_i, & \text{otherwise}.
\end{cases}
\label{eq:app_action_advantage}
\end{equation}
For a positive trajectory, the allocation preserves
$\sum_{t\in\mathcal I_i}L_{i,t}\widetilde A_{i,t}=A_iT_i$.
This is conservation of advantage coefficients, not of the loss,
gradient norm, or trajectory-length-normalized weight.
Nonpositive advantages remain unchanged across the trajectory;
weighting negative advantages by local progress could otherwise
penalize improving actions more strongly. Retrieval, submission,
and other retained response segments outside $\mathcal I_i$ also
retain their original trajectory advantage.

\clearpage
\subsection{Response Masks and Policy Optimization}
\label{app:rl_optimization}
\AppBookmark{1}{app:rl_optimization}{\thesubsection\space Response Masks and Policy Optimization}

The response mask selects every sampled token in an action response.
For a valid response, training therefore covers the full action JSON,
including the image choice, target and preserve sets, instruction,
and structural tokens. The protocol uses direct action JSON without
a separate reasoning channel or reasoning mask.
If extra text makes a response invalid, its retained sampled tokens
can still receive the invalid trajectory's training signal.
History, tool results, and verifier feedback are conditioning inputs,
not sampled response targets.

Let $\mathcal B$ be the trajectories retained for an optimizer step,
$\mathcal T_i$ the mask-selected response tokens in trajectory $i$,
and $\widetilde A_{i,\ell}$ the advantage of the action containing
token $y_{i,\ell}$. For its multimodal prefix $H_{i,\ell}$, define
\begin{equation}
\begin{aligned}
\rho_{i,\ell}
&=\frac{\pi_\theta(y_{i,\ell}\mid H_{i,\ell})}
{\pi_{\mathrm{old}}(y_{i,\ell}\mid H_{i,\ell})},\\
d_{i,\ell}
&=\log\pi_{\mathrm{ref}}(y_{i,\ell}\mid H_{i,\ell})
-\log\pi_\theta(y_{i,\ell}\mid H_{i,\ell}),\\
\mathcal K_{i,\ell}&=\exp(d_{i,\ell})-d_{i,\ell}-1.
\end{aligned}
\label{eq:app_kl_estimator}
\end{equation}
This is the configured low-variance KL estimator. The corresponding minimization loss is
\begin{equation}
\begin{aligned}
\mathcal L(\theta)
={}&\frac{1}{|\mathcal B|}\sum_{i\in\mathcal B}
\sum_{\ell\in\mathcal T_i}
\Bigl[-\min\bigl\{\rho_{i,\ell}\widetilde A_{i,\ell},\\[-0.2em]
&\qquad \operatorname{clip}(\rho_{i,\ell},0.8,1.28)
\widetilde A_{i,\ell}\bigr\}
+0.02\mathcal K_{i,\ell}\Bigr].
\end{aligned}
\label{eq:app_token_sum_loss}
\end{equation}
The loss uses trajectory-mean, token-sum reduction: tokens are summed
within trajectories, then trajectories are averaged. There is no
division by individual trajectory length or the total token count.
Group statistics use the four returns within each prompt group;
$|\mathcal B|$ instead counts the retained trajectories assigned
to an optimizer step, potentially spanning several prompt groups.
Longer responses therefore contribute more token-level terms;
NACA's coefficient conservation does not remove this length effect.

The trainer processes one action segment at a time and accumulates
all segments assigned to an optimizer step. Multiplying local loss
sums by the world size divided by the retained trajectory count,
followed by FSDP gradient averaging, gives the reduction in
Eq.~\eqref{eq:app_token_sum_loss}. Accumulation is determined by the
number of segments, not a fixed microstep count.

\begin{appbox}[appGreen]{Zero action credit is not a loss mask}
NACA changes the advantage but not the response mask. Under normal
positive-credit allocation, an image action with $w_{i,t}=0$ has no
positive policy-gradient term but retains its KL term. KL uses the
same response mask for generation, editing, retrieval, submission,
and retained invalid responses. This differs from a zero-variance
group, which is removed before optimization and contributes no KL.
\end{appbox}
\begin{table}[!htbp]
\centering
\small
\caption{Credit and KL for responses in retained optimizer data.
The first two rows assume effective positive progress is available.}
\label{tab:rl_credit_and_kl}
\setlength{\tabcolsep}{5pt}
\renewcommand{\arraystretch}{1.12}
\begin{tabularx}{\linewidth}{@{}X p{0.22\linewidth}p{0.13\linewidth}@{}}
\toprule
\rowcolor{appTableHead}
\textbf{Response or input} & \textbf{Advantage} & \textbf{KL term}\\
\midrule
Eligible image action, $A_i>0$, $q_{i,t}>\delta$ & $A_iw_{i,t}$ & Retained\\
Eligible image action, $A_i>0$, $q_{i,t}\leq\delta$ & $0$ & Retained\\
Eligible image action, positive-trajectory fallback & $A_i$ & Retained\\
Image action in a nonpositive-advantage trajectory & $A_i$ & Retained\\
Retrieval, submission, or retained invalid response & $A_i$ & Retained\\
Context, tool results, and verifier feedback & Not a target & Excluded\\
Zero-variance group & Group excluded & Excluded\\
\bottomrule
\end{tabularx}
\end{table}

\clearpage
\section{Optimization Potential of Historical Continuation States}
\label{app:historical_theory}
\AppBookmark{0}{app:historical_theory}{\thesection\space Optimization Potential of Historical Continuation States}
Historical-image access changes the set of continuations available to a policy. The result below concerns that feasible set and the optimal reachable value, not the performance of a particular learned checkpoint. It preserves the same terminal utility and controlled comparison as the main formulation.

\subsection{Interaction State and Continuation State}
\label{app:joint_state_action}
\AppBookmark{1}{app:joint_state_action}{\thesubsection\space Interaction State and Continuation State}
The interaction state $s_t$ contains the task, previous actions and observations, available images, and remaining budget. A continuation state is one image $I_{k_t}$ chosen as the input to an edit. The policy produces
\begin{equation}
a_t=(o_t,\eta_t)\sim\pi_\theta(\cdot\mid s_t),
\qquad \eta_t=(k_t,u_t)\quad\text{when }o_t=\textsc{Edit},
\label{eq:app_joint_action}
\end{equation}
where $o_t$ is one of generation, editing, retrieval, and submission, and $u_t$ contains the edit specification. The image choice and edit specification are components of the same decision. Selecting an earlier image does not remove observations obtained after its creation or restore spent image calls. Generation creates a new root; editing creates a child of the selected candidate.

\subsection{Feasible-Set Monotonicity}
\label{app:source_monotonicity}
\AppBookmark{1}{app:source_monotonicity}{\thesubsection\space Feasible-Set Monotonicity}
Consider a decision state with interaction history $H_t$ and $b$ remaining image calls. A source-access rule $\Gamma$ maps each reachable history $H$ to the set $\Gamma(H)$ of images that may be selected as editing inputs. Hold the prompt, observations, frozen generation and editing backends, non-edit actions, submission set, and image-call accounting fixed. Let $\Pi_\Gamma$ denote the policies respecting this access rule. Using $U(I)=P(I)+0.25g(I)$, define
\begin{equation}
V^\star_\Gamma(H_t,b)
=\sup_{\pi\in\Pi_\Gamma}
\mathbb E_\pi[U(I_{\mathrm{sub}})\mid H_t,b].
\label{eq:app_reachable_value}
\end{equation}

\begin{appbox}[appViolet]{Proposition 1. Monotonicity under source-set expansion}
Suppose $\Gamma_1(H)\subseteq\Gamma_2(H)$ for every history reachable from $(H_t,b)$. Then
\begin{equation}
V^\star_{\Gamma_2}(H_t,b)\geq V^\star_{\Gamma_1}(H_t,b).
\label{eq:app_source_monotonicity}
\end{equation}
\end{appbox}

\paragraph{Proof.}
Every policy feasible under $\Gamma_1$ is also feasible under $\Gamma_2$: the expanded rule adds editing choices but removes none of the actions available to the restricted policy. Hence $\Pi_{\Gamma_1}\subseteq\Pi_{\Gamma_2}$, and optimizing the same terminal objective over the larger policy class cannot decrease its value. \hfill$\square$

\subsection{Interpretation and Scope}
\label{app:theory_scope}
\AppBookmark{1}{app:theory_scope}{\thesubsection\space Interpretation and Scope}
The inclusion must hold over reachable histories, not just at the initial branch point. A comparison that also changes observations, available generation actions, submission rules, or budgets is outside the stated controlled setting. Even when the inclusion holds, a trained policy need not realize the larger class's optimal value. The proposition neither guarantees strict improvement nor requires the planner to reject the current best image.

Current quality and continuation potential are distinct. Two candidates can preserve different useful content and require different edits, even when one currently has a lower score. Appendix~\ref{app:quality_potential_case} gives an observed successful continuation from a lower-scoring candidate. Appendix~\ref{app:joint_best_case} reports an executed Joint/Best-only pair from a common prefix. The former is an observation about one trajectory; the latter compares two realized continuations. Neither is a proof that every edit from the higher-scoring image would fail.

\section{Analysis Protocols and Additional Results}
\label{app:analysis}
\subsection{Controlled Teacher Continuations}
\label{app:teacher_continuations}
To quantify the benefit of joint selection, we select 200 prefixes from a pool of 800 teacher trajectories where the task remains unsolved after three image calls. Each prefix provides a shared starting history for Joint, Latest-only, and Best-only, with two image calls remaining. All three use the same teacher and interaction protocol, differing only in edit-source access. Latest-only uses the latest image, Best-only uses the current highest-GM image, and Joint selects the source together with its editing instruction. Joint achieves a mean final GM of 55.2, compared with 52.4 for Best-only and 50.2 for Latest-only. Its improvement over Best-only is 2.8 points (paired 95\% CI: [0.46, 5.15]). Joint also increases the proportion of prefixes producing a new candidate above the shared prefix best from 57.0\% with Best-only to 65.0\%. These results support the value of joint selection beyond fixing the editing source to the current highest-scoring image.

\subsection{Learning from Joint Demonstrations}
\label{app:joint_demonstrations}
A second experiment examines whether joint demonstrations lead to better performance after SFT. An additional 800 Best-only teacher trajectories are collected using the same prompts and initial images as the Joint trajectories. The two demonstration sets are used to train separate student policies via SFT. Both policies are then evaluated on the official GenEval2 evaluation set of 800 prompts under the same evaluation protocol used for our main results. The student trained on Joint demonstrations achieves 81.5 GM, compared with 77.5 for its Best-only-trained counterpart, an improvement of 4.0 points (paired 95\% CI: [2.81, 5.18]). The advantage is thus observed not only in controlled teacher continuations but also in the benchmark performance of the trained policies.

\subsection{Historical-Source Access after Regression}
\label{app:student_continuations}
We additionally evaluate historical source access in a separate Student experiment on 240 degraded prefixes, where the latest candidate is strictly worse than the historical best. Here, success requires generating a new candidate that surpasses the prefix best; simply returning an existing historical image does not count. Allowing historical edit sources increases this discovery rate from 60.00\% to 71.67\%, an improvement of 11.67 percentage points (paired 95\% CI: [5.00, 18.33], \(p=0.00109\)).

The two arms inherit the same executed prefix and remaining image-call budget. Full-history can edit completed historical candidates; Latest-only can edit only the latest candidate. The 240 selected prefixes consist of 143 with two completed images and 97 with three. Both arms execute fresh suffixes. Discovery compares newly generated images with the frozen prefix best; selecting an already existing candidate does not count as discovery. The archived study ranks candidates by passed-constraint count followed by GM, rather than substituting the GM-first control of the current Joint interface.

The discovery counts are 172/240 and 144/240. There are 49 pairs favoring Full-history, 21 favoring Latest-only, and 170 ties. The paired 95\% confidence interval is [5.00, 18.33] percentage points, and the two-sided exact paired test gives $p=0.00109$. Mean executed suffix calls are 2.233 and 2.104: the matched quantity is the remaining budget upper bound. All 365 historical edits in the Full-history arm select the then-current best. This comparison therefore measures historical-source access versus latest-only access; the teacher study above separately compares Joint with Best-only.

\subsection{Cross-Benchmark Category Results}
\label{app:cross_benchmark_details}
Tables~\ref{tab:app-geneval-details} and~\ref{tab:app-genevalpp-details} give the category results corresponding to the main-paper overall scores, 0.9844 on GenEval and 0.8929 on GenEval++. Scores here use the 0--1 scale; the main cross-benchmark table displays percentages.
\begingroup
\definecolor{oodband}{gray}{0.95}
\definecolor{oodours}{gray}{0.92}
\newcommand{\oodbest}[1]{\textbf{#1}}
\newcommand{\oodsecond}[1]{#1}
\setlength{\tabcolsep}{0pt}
\renewcommand{\arraystretch}{1.30}

\begin{table}[!t]
\centering
\caption{Cross-benchmark generalization. Bold scores indicate the best result in each column.}
\label{tab:app-geneval-details}

\resizebox{\textwidth}{!}{%
\begin{minipage}{230mm}
\fontsize{11}{13}\selectfont
\begin{tabular}{@{}>{\hspace{7pt}\raggedright\arraybackslash}p{0.30\linewidth}@{}*{7}{>{\centering\arraybackslash}p{0.10\linewidth}@{}}}
\toprule
\multicolumn{8}{c}{\textbf{GenEval}} \\
\cmidrule(lr){1-8}
\textbf{Method}
  & Single obj.$\uparrow$ & Two obj.$\uparrow$ & Counting$\uparrow$
  & Colors$\uparrow$ & Position$\uparrow$ & Color attr.$\uparrow$
  & Overall$\uparrow$ \\
\midrule
\rowcolor{oodband}\multicolumn{8}{@{\hspace{7pt}}l}{\textsc{Direct generation}} \\
FLUX.1-dev~\citep{zhuo2025reflectionflow}
  & \oodsecond{0.9900} & 0.8100 & 0.7500 & 0.8000
  & 0.2100 & 0.4800 & 0.6700 \\
Qwen-Image~\citep{wu2025qwenimage}
  & \oodsecond{0.9900} & 0.9200 & 0.8900 & 0.8800
  & 0.7600 & 0.7700 & 0.8700 \\
\addlinespace[2pt]
\rowcolor{oodband}\multicolumn{8}{@{\hspace{7pt}}l}{\textsc{Agentic generation and refinement}} \\
T2I-Copilot$^{\dagger}$~\citep{chen2025t2icopilot,jiang2026raise}
  & \oodsecond{0.9900} & 0.9100 & 0.6800 & 0.8600
  & 0.5500 & 0.4600 & 0.7400 \\
GEMS (Z-Image-Turbo)~\citep{he2026gems}
  & \oodsecond{0.9900} & 0.9200 & 0.9100 & 0.8800
  & 0.8300 & 0.6500 & 0.8600 \\
FiRe-GRPO (3rd round)~\citep{kim2026fimr}
  & \oodsecond{0.9900} & 0.9400 & 0.7200 & 0.9300
  & 0.8500 & 0.7700 & 0.8700 \\
Generation Navigator~\citep{liu2026navigator}
  & -- & -- & -- & --
  & -- & -- & 0.8843 \\
ReflectionFlow~\citep{zhuo2025reflectionflow}
  & \oodbest{1.0000} & 0.9800 & 0.9000 & 0.9600
  & 0.9300 & 0.7200 & 0.9100 \\
RAISE~\citep{jiang2026raise}
  & \oodbest{1.0000} & \oodbest{1.0000} & 0.9500 & \oodbest{0.9800}
  & 0.8300 & 0.8700 & 0.9400 \\
coDrawAgents~\citep{li2026codrawagents}
  & \oodbest{1.0000} & 0.9600 & 0.9400 & 0.9700
  & 0.9500 & 0.8100 & 0.9400 \\
CoIG~\citep{kim2025coig}
  & 0.9875 & 0.9899 & 0.8625 & 0.9255
  & \oodsecond{0.9900} & \oodsecond{0.9350} & 0.9484 \\
OctoT2I~\citep{jiang2026octot2i}
  & \oodbest{1.0000} & \oodsecond{0.9900} & 0.9500 & 0.9400
  & \oodbest{1.0000} & 0.8600 & 0.9600 \\
\addlinespace[2pt]
\rowcolor{oodours}
\textbf{GenRevisit}
  & \oodbest{1.0000} & 0.9899 & \oodbest{0.9875} & \oodsecond{0.9787}
  & 0.9800 & \oodbest{0.9700} & \oodbest{0.9844} \\
\bottomrule
\end{tabular}
\end{minipage}%
}
\end{table}
\endgroup

\begingroup
\hypersetup{hidelinks}
\begin{table}[!t]
\centering
\caption{Comparison on GenEval++. The seven categories follow the benchmark taxonomy. Higher is better; bold marks the highest displayed result in each column.}
\label{tab:app-genevalpp-details}
\begingroup
\definecolor{surveyband}{HTML}{F3F3F3}
\definecolor{surveyours}{HTML}{EFEFEF}
\fontsize{11.5}{13.5}\selectfont
\setlength{\tabcolsep}{0pt}
\renewcommand{\arraystretch}{1.22}
\resizebox{\textwidth}{!}{%
\begin{minipage}{240mm}
\begin{tabular}{@{}>{\hspace{3pt}\raggedright\arraybackslash}p{64mm}>{\raggedright\arraybackslash}p{40mm}*{8}{>{\centering\arraybackslash}p{17mm}}@{}}
\toprule[0.8pt]
\multirow{3}{*}{\textbf{Method}} & \multirow{3}{*}{\textbf{Image backend}} & \multicolumn{7}{c}{\textbf{Category-wise accuracy}} & \multirow{3}{*}{\textbf{Overall}} \\
\cmidrule(lr){3-9}
 & & \textbf{Color} & \textbf{Count} & \shortstack{\textbf{Color/}\\\textbf{Count}} & \shortstack{\textbf{Color/}\\\textbf{Pos}} & \shortstack{\textbf{Pos/}\\\textbf{Count}} & \shortstack{\textbf{Pos/}\\\textbf{Size}} & \shortstack{\textbf{Multi-}\\\textbf{Count}} & \\
\midrule[0.45pt]
\rowcolor{surveyband}\multicolumn{10}{@{}l}{\rule{0pt}{12pt}\textsc{Direct generation}} \\
\mbox{FLUX.1-dev~\citep{jiang2026genagent}} & FLUX.1-dev & 0.4000 & 0.6000 & 0.2500 & 0.2500 & 0.0750 & 0.4000 & 0.3000 & 0.3250 \\
\mbox{BAGEL~\citep{ye2025echo4o}} & BAGEL & 0.3250 & 0.6000 & 0.2500 & 0.3250 & 0.2500 & 0.4750 & 0.3750 & 0.3710 \\
\mbox{Qwen-Image~\citep{jiang2026genagent}} & Qwen-Image & 0.8750 & 0.7250 & 0.7250 & 0.6000 & 0.4750 & 0.7250 & 0.5500 & 0.6680 \\
\mbox{Echo-4o~\citep{ye2025echo4o}} & BAGEL & 0.8000 & 0.5750 & 0.5500 & 0.7750 & 0.6250 & 0.8000 & 0.6250 & 0.6790 \\
\rowcolor{surveyband}\multicolumn{10}{@{}l}{\rule{0pt}{12pt}\textsc{Closed-source generation}} \\
\mbox{GPT-4o~\citep{ye2025echo4o}} & GPT-4o & 0.9000 & 0.6750 & 0.7250 & 0.6250 & 0.6000 & 0.8000 & 0.8500 & 0.7390 \\
\mbox{GPT-Image-1.5~\citep{ye2026genclaw}} & GPT-Image-1.5 & 0.8500 & 0.9500 & 0.7750 & 0.5000 & 0.6000 & 0.7750 & 0.8000 & 0.7500 \\
\mbox{Gemini-3.1 Flash-Image~\citep{ye2026genclaw}} & Gemini-3.1 & 0.8500 & 0.8500 & 0.8000 & 0.7500 & 0.6750 & 0.7250 & 0.7750 & 0.7750 \\
\rowcolor{surveyband}\multicolumn{10}{@{}l}{\rule{0pt}{12pt}\textsc{Agentic generation and refinement}} \\
\mbox{PromptEnhancer~\citep{jiang2026genagent}} & FLUX.1-dev & 0.5000 & 0.6250 & 0.2250 & 0.3750 & 0.1250 & 0.4500 & 0.3750 & 0.3820 \\
\mbox{ReflectionFlow~\citep{jiang2026genagent}} & FLUX.1-dev & 0.4000 & 0.6250 & 0.2750 & 0.2750 & 0.2000 & 0.4250 & 0.3250 & 0.3610 \\
\mbox{FiRe (3rd round)~\citep{kim2026fimr}} & Janus-Pro-7B & 0.7750 & 0.4880 & 0.4500 & 0.5690 & 0.2810 & 0.3560 & 0.4880 & 0.4910 \\
\mbox{GenAgent~\citep{jiang2026genagent}} & FLUX.1-dev & 0.7500 & 0.6750 & 0.3750 & 0.5250 & 0.4500 & 0.6250 & 0.5250 & 0.5610 \\
\mbox{GenAgent~\citep{jiang2026genagent}} & Qwen-Image & 0.7750 & 0.7750 & 0.6500 & 0.8000 & 0.6000 & 0.7250 & 0.7500 & 0.7250 \\
\mbox{Mind-Brush~\citep{he2026mindbrush}} & Qwen-Image & 0.7750 & 0.7000 & 0.7750 & 0.7500 & 0.8500 & 0.7750 & 0.8500 & 0.7820 \\
\mbox{GenClaw~\citep{ye2026genclaw}} & Gemini-3.1 & 0.8750 & 0.9500 & 0.8250 & \textbf{0.8500} & \textbf{0.9250} & 0.8250 & \textbf{0.9000} & 0.8786$^{*}$ \\
\rowcolor{surveyband}\multicolumn{10}{@{}l}{\rule{0pt}{12pt}\textsc{Our method}} \\
\rowcolor{surveyours}
\textbf{GenRevisit} & Qwen-Image-2512 & \textbf{0.9500} & \textbf{1.0000} & \textbf{0.9000} & \textbf{0.8500} & 0.8250 & \textbf{0.8750} & 0.8500 & \textbf{0.8929} \\
\bottomrule[0.8pt]
\end{tabular}
\end{minipage}%
}
\endgroup
\vspace{4pt}
\begin{minipage}{\linewidth}
\fontsize{8}{9.5}\selectfont
\end{minipage}
\end{table}
\endgroup

\clearpage

\clearpage
\section{Case Studies and Qualitative Comparisons}
\label{app:cases}
\AppBookmark{0}{app:cases}{\thesection\space Case Studies and Qualitative Comparisons}
The recorded cases below distinguish teacher continuations, student behavior, and training-credit diagnostics. GM is shown on a 0--100 scale; passed/$K$ reports saved verifier judgments. Arrows show executed edits, while separate submission labels identify the final output.

\subsection{Joint and Best-only Continuations}
\label{app:joint_best_case}
\AppBookmark{1}{app:joint_best_case}{\thesubsection\space Joint and Best-only Continuations}
For ``six green yellow flamingos,'' two GPT-5.5 teacher continuations share three evaluated images and two remaining calls. Both initially edit the prefix GM-best. Joint's last edit instead uses \texttt{a\_002}, now neither latest nor GM-best, and adds green while retaining yellow, count, and layout. Its submitted image reaches GM 94.81 and 4/4 passed constraints. Best-only submits the existing \texttt{a\_001} at GM 82.65 and 3/4.

\begin{figure}[H]
\centering
\input{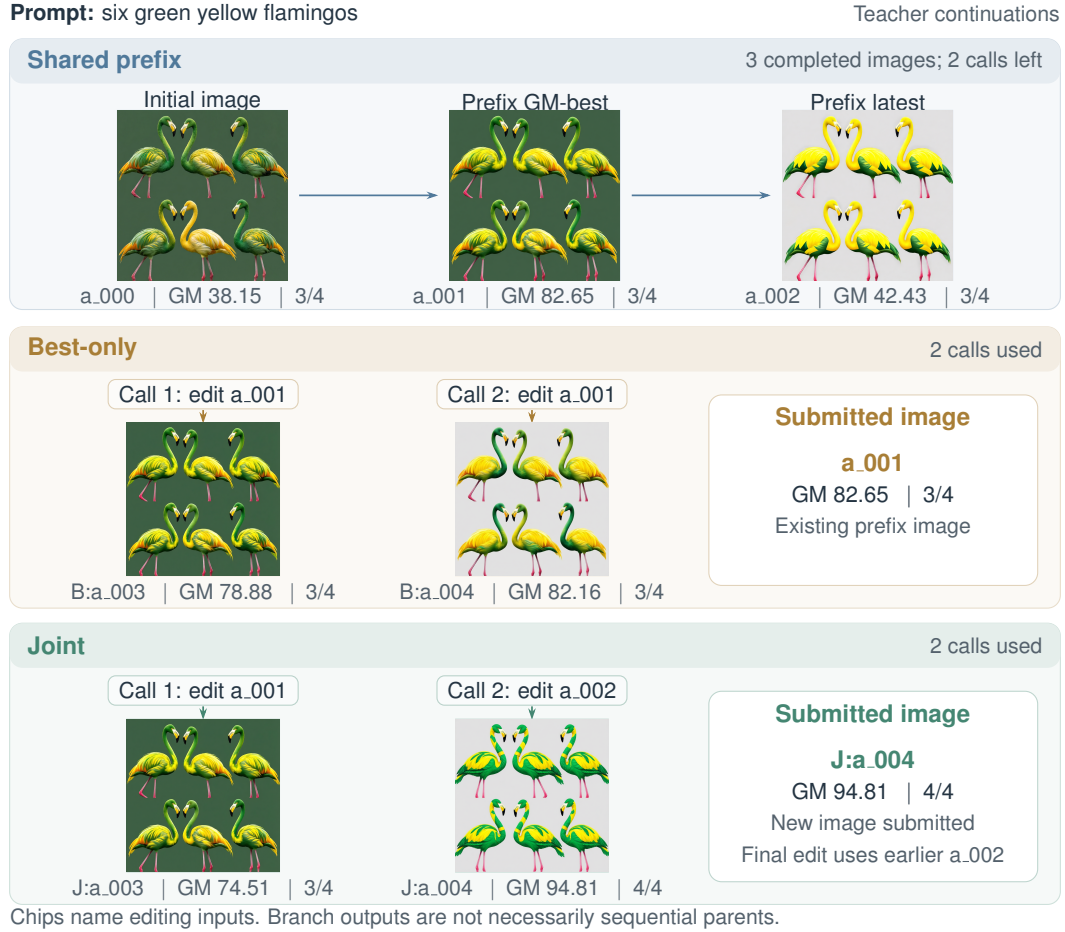}
\caption{\textbf{Joint and Best-only from a shared prefix.} Source labels identify the actual editing inputs. Joint's final parent is \texttt{a\_002}, not \texttt{J:a\_003}; Best-only submits shared \texttt{a\_001}, not \texttt{B:a\_004}. Suffix identifiers are branch-local. The final Joint instruction appears in Appendix~\ref{app:recorded_instruction}.}
\label{fig:flamingo_pair}
\end{figure}

This selected teacher pair does not establish overall or student-policy superiority. The separate Latest-only run on this prompt scores higher than Joint; the figure is specifically a Joint/Best-only comparison. Aggregate analysis is reserved for Appendix~\ref{app:analysis}.

\clearpage
\subsection{Historical Continuation and Latest-only Refinement}
\label{app:historical_latest_case}
\begin{figure}[H]
\centering
\input{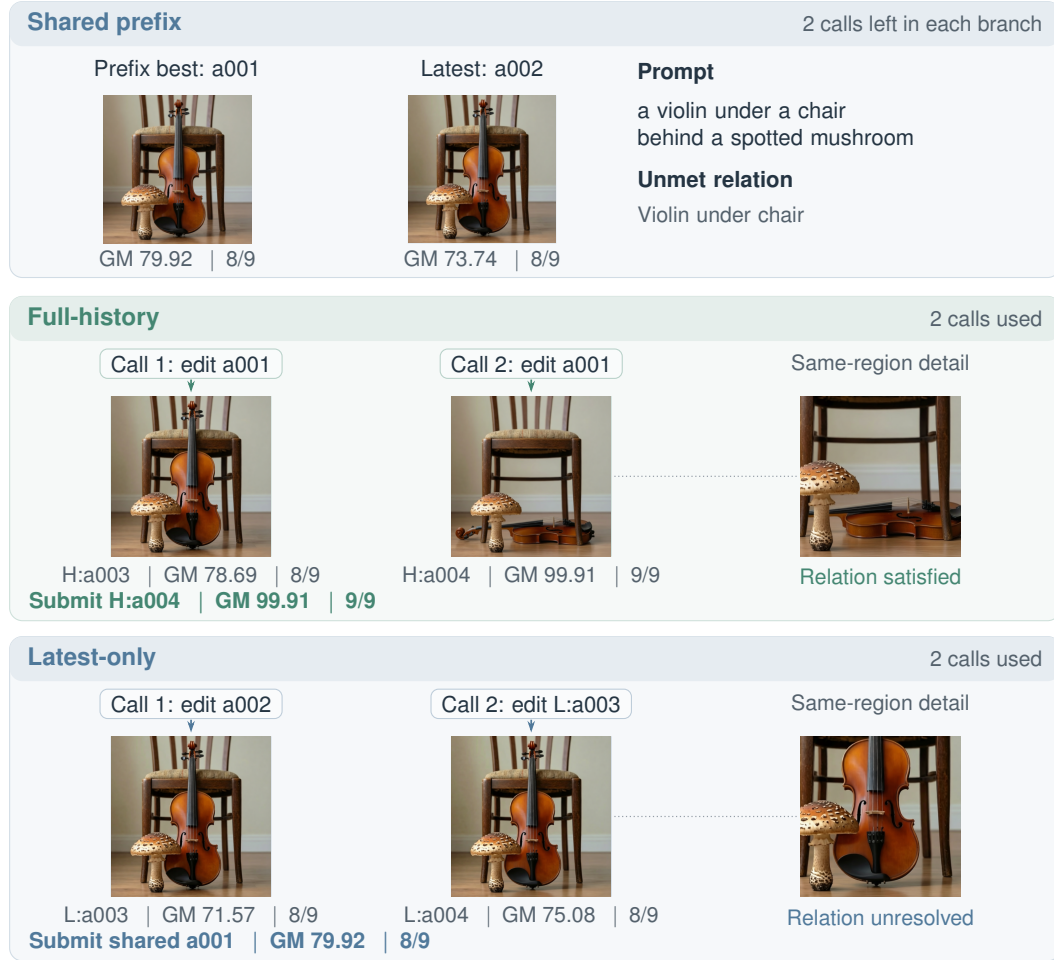}
\caption{\textbf{An archived historical-access versus latest-only pair.} Both branches use two additional calls. Full-history has parent relations \texttt{a001}$\to$\texttt{H:a003} and \texttt{a001}$\to$\texttt{H:a004}; Latest-only follows \texttt{a002}$\to$\texttt{L:a003}$\to$\texttt{L:a004}. The detail panels reproduce the same-region crops from the archived figure and refer to the last generated images, not necessarily to the submissions. Dotted lines denote detail views, not edits.}
\label{fig:violin_pair}
\end{figure}

\AppBookmark{1}{app:historical_latest_case}{\thesubsection\space Historical Continuation and Latest-only Refinement}
The archived student comparison in Figure~\ref{fig:violin_pair} uses the prompt ``a violin under a chair behind a spotted mushroom.'' The prefix best \texttt{a001} and latest \texttt{a002} both pass eight of nine constraints. The remaining failed relation is the violin being under the chair. Both branches have exactly two image calls left.

Full-history edits \texttt{a001} on both calls. The first result remains incomplete, while the second places the violin horizontally under the chair and passes all nine constraints (GM 99.91). Latest-only edits \texttt{a002} and then its own new result. Its final generated image still fails the relation, so the branch submits the original prefix best (GM 79.92). Thus, the last generated image and the submitted image are different in the Latest-only branch.

This case concerns access to historical editing inputs relative to latest-only editing. It is not relabelled as a Joint/Best-only experiment: the historical edits shown here both use the current best. Its complete statistical protocol is separate from the present case description.

\clearpage
\subsection{Current Quality and Continuation Potential}
\label{app:quality_potential_case}
\AppBookmark{1}{app:quality_potential_case}{\thesubsection\space Current Quality and Continuation Potential}
An earlier teacher trajectory for ``three cats chasing five sheeps'' contains two candidates with complementary satisfied constraints. The current best \texttt{a000} passes the sheep-count requirement but fails the chasing relation. Candidate \texttt{a001} passes the chasing relation but fails sheep counting. Both pass four of five constraints, although \texttt{a001} has lower GM (35.68 versus 61.88).

\begin{figure}[H]
\centering
\input{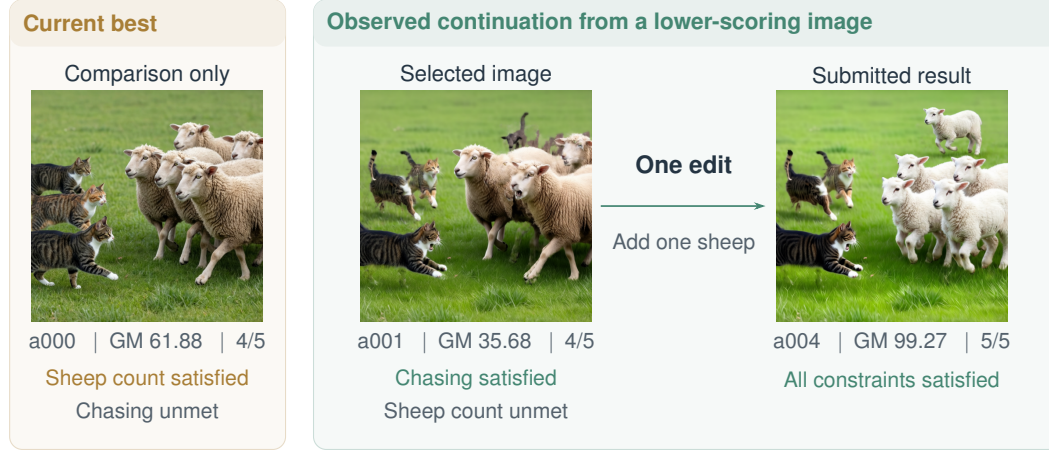}
\caption{\textbf{A successful continuation from below the current best.} The only executed edge shown is \texttt{a001}$\to$\texttt{a004}. \texttt{a000} is a comparison candidate, not an intermediate image in that edit. Discrete labels report the saved verifier outcomes. This is one observed trajectory rather than a paired continuation from both possible starting images.}
\label{fig:cats_continuation}
\end{figure}

With one call remaining, the teacher edits \texttt{a001} to add a separately countable sheep while preserving the chasing relation. The submitted result \texttt{a004} passes all five constraints with GM 99.27. Figure~\ref{fig:cats_continuation} places the comparison candidate separately from the executed editing path.

The observation establishes that the successful final edit did not have to start from the highest-scoring available image. It does not establish that every possible edit from \texttt{a000} would fail, or compare the two candidates' expected continuation values. This distinction is consistent with the scope of Proposition~1 in Appendix~\ref{app:source_monotonicity}.

\clearpage
\subsection{Action Credit on a Recorded Trajectory}
\label{app:credit_case}
\AppBookmark{1}{app:credit_case}{\thesubsection\space Action Credit on a Recorded Trajectory}
The training trajectory for ``a purple elephant chasing a zebra'' has positive group-relative advantage $A_i=1.730187$. Table~\ref{tab:credit_case} reports the stored local progress and NACA coefficients. The GRPO reference is the uniform broadcast coefficient on this same sampled trajectory, not a separately sampled GRPO run.

\begin{table}[H]
\centering\small
\caption{Recorded image-action credit. GM is on a 0--100 scale. The printed coefficients are rounded; token-weighted conservation applies to the stored full-precision values.}
\label{tab:credit_case}
\setlength{\tabcolsep}{2.8pt}
\begin{tabularx}{\linewidth}{@{}lXrrrrr@{}}
\toprule
\rowcolor{appTableHead}
\textbf{Image action} & \textbf{Pass; GM} & $L_{i,t}$ & $q_{i,t}$ & $w_{i,t}$ & \textbf{GRPO} & \textbf{NACA}\\
\midrule
Generate & 5/6; 62.03 & 218 & .824812 & 1.898143 & 1.730187 & 3.284142\\
$\mathrm{a000}\!\to\!\mathrm{a001}$ & 5/6; 82.54 & 342 & .046820 & .107747 & 1.730187 & .186422\\
$\mathrm{a001}\!\to\!\mathrm{a002}$ & 5/6; 54.16 & 376 & 0 & 0 & 1.730187 & 0\\
$\mathrm{a001}\!\to\!\mathrm{a003}$ & 6/6; 95.64 & 380 & .989549 & 2.277251 & 1.730187 & 3.940071\\
\bottomrule
\end{tabularx}
\end{table}

\begin{figure}[H]
\centering
\begin{tikzpicture}[appfigure]
\path[use as bounding box] (0,0) rectangle (13.9,6.6);
\path[apppanel,fill=appBlue!5!white,draw=appBlue!26!white] (0,0) rectangle (13.9,1.12);
\path[fill=appBlue!14!white,rounded corners=1.7mm] (0,0) rectangle (13.9,0.59);
\node[anchor=west,appfigtitle,text=appBlue] at (0.23,0.295) {One recorded trajectory};
\node[anchor=east,appmeta] at (13.67,0.295) {$A_i=1.730187$};
\node[anchor=west,appmeta] at (0.23,0.85) {a purple elephant chasing a zebra};
\draw[draw=appRule,line width=.35pt] (1.05,5.64) -- (13.62,5.64);
\node[anchor=east,appmeta] at (0.83,5.64) {0};
\draw[draw=appRule,line width=.35pt] (1.05,4.84) -- (13.62,4.84);
\node[anchor=east,appmeta] at (0.83,4.84) {1};
\draw[draw=appRule,line width=.35pt] (1.05,4.039999999999999) -- (13.62,4.039999999999999);
\node[anchor=east,appmeta] at (0.83,4.039999999999999) {2};
\draw[draw=appRule,line width=.35pt] (1.05,3.2399999999999993) -- (13.62,3.2399999999999993);
\node[anchor=east,appmeta] at (0.83,3.2399999999999993) {3};
\draw[draw=appRule,line width=.35pt] (1.05,2.4399999999999995) -- (13.62,2.4399999999999995);
\node[anchor=east,appmeta] at (0.83,2.4399999999999995) {4};
\node[anchor=west,appmeta] at (0,1.63) {Action advantage};
\node[anchor=west,appmeta] at (8.0,1.61) {Solid: NACA};
\node[anchor=west,appmeta] at (10.66,1.61) {Dashed: GRPO};
\path[fill=appGreen!80!white,draw=appGreen,line width=.45pt] (1.8200000000000003,3.0126863999999993) rectangle (3.08,5.64);
\node[anchor=center,appfigtitle] at (2.45,2.8026863999999994) {3.28};
\path[fill=appGreen!80!white,draw=appGreen,line width=.45pt] (4.89,5.490862399999999) rectangle (6.1499999999999995,5.64);
\node[anchor=center,appfigtitle] at (5.52,5.280862399999999) {0.19};
\draw[appAmber,line width=1.2pt] (7.96,5.64) -- (9.22,5.64);
\node[anchor=center,appfigtitle] at (8.59,5.43) {0};
\path[fill=appGreen!80!white,draw=appGreen,line width=.45pt] (11.03,2.4879431999999992) rectangle (12.290000000000001,5.64);
\node[anchor=center,appfigtitle] at (11.66,2.2779431999999993) {3.94};
\draw[appBlue,dashed,line width=.75pt] (1.05,4.2558504) -- (13.62,4.2558504);
\node[anchor=east,appmeta,text=appBlue] at (13.55,4.0858504) {1.73};
\node[anchor=center,appfigtitle] at (2.45,5.98) {Generate};
\node[anchor=center,appmeta] at (2.45,6.37) {$\varnothing$\,$\to$\,a000};
\node[anchor=center,appfigtitle] at (5.52,5.98) {Edit};
\node[anchor=center,appmeta] at (5.52,6.37) {a000\,$\to$\,a001};
\node[anchor=center,appfigtitle] at (8.59,5.98) {Edit};
\node[anchor=center,appmeta] at (8.59,6.37) {a001\,$\to$\,a002};
\node[anchor=center,appfigtitle] at (11.66,5.98) {Revisit};
\node[anchor=center,appmeta] at (11.66,6.37) {a001\,$\to$\,a003};
\end{tikzpicture}
\caption{\textbf{NACA changes action credit within a single positive-advantage rollout.} Solid bars show NACA action advantages; the dashed line is the common GRPO broadcast value. Each bar represents an image action, not a token count. The zero bar removes positive policy-gradient credit, not the response's KL term. Retrieval and submission are not shown and keep their trajectory advantage.}
\label{fig:credit_case}
\end{figure}
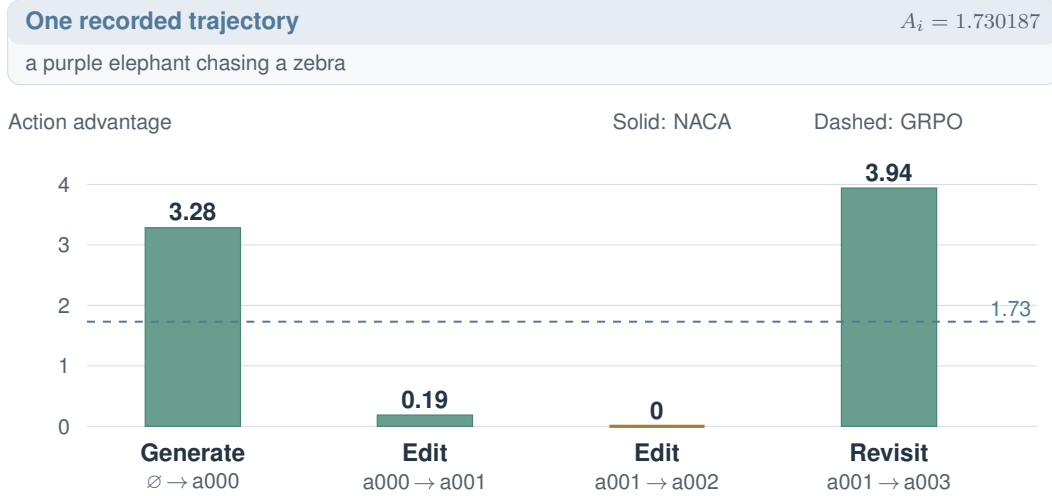

The edit from \texttt{a001} to \texttt{a002} retains five passed constraints but reduces GM, so its utility does not increase and its positive action advantage is zero. A small GM-only improvement earlier in the trajectory receives positive but reduced credit. The final edit revisits \texttt{a001}, reaches six passed constraints, and receives the largest multiplier. Figure~\ref{fig:credit_case} shows the coefficients on one shared scale.

The four image actions contain 1,316 trainable tokens in total. The recorded multipliers preserve $\sum_t L_{i,t}w_{i,t}=1,316$ up to floating-point precision. This preserves advantage coefficients rather than a gradient norm. The case illustrates the allocation mechanism; it is not a new estimate of the performance difference between separately trained GRPO and NACA policies.

\clearpage
\subsection{Additional Qualitative Outputs}
\label{app:qualitative}
\subsubsection{Designed-Prompt Results}
\begin{figure}[H]
  \centering
  \includegraphics[width=\linewidth,height=0.78\textheight,keepaspectratio]{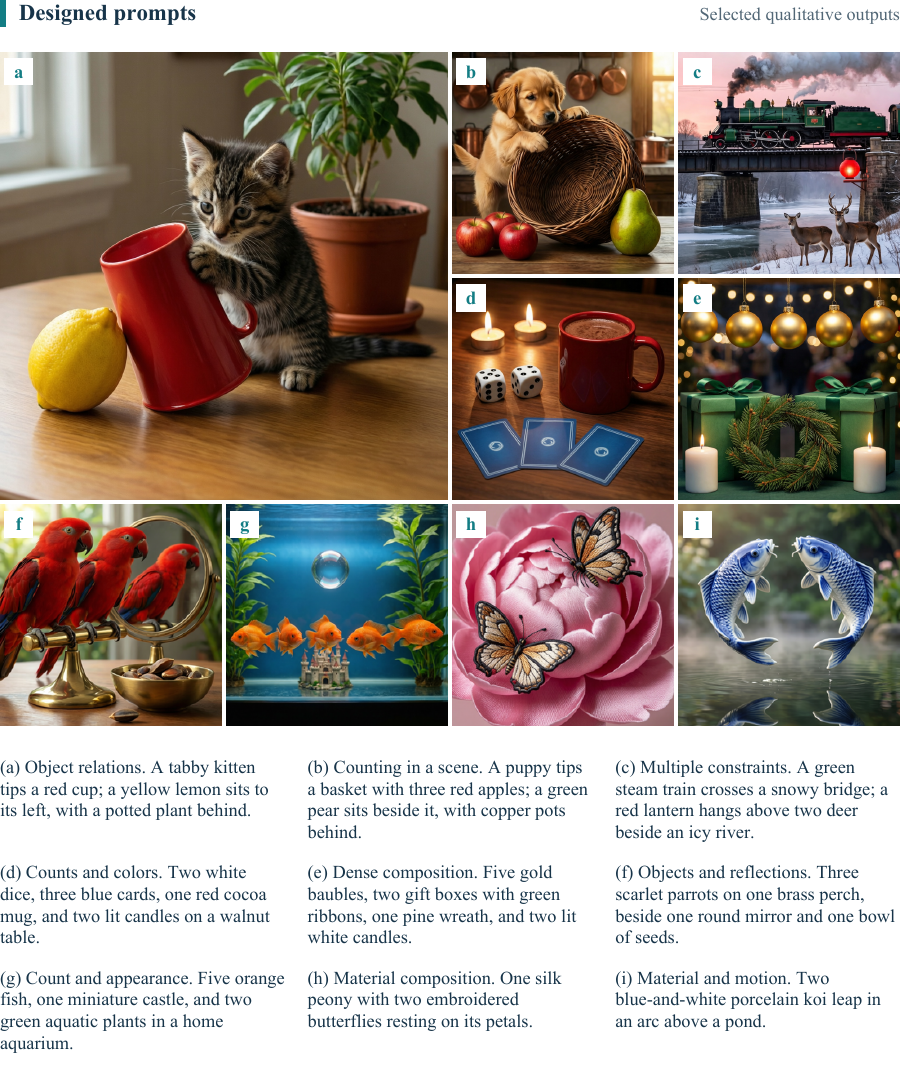}
  \caption{\textbf{Selected outputs on designed prompts.}
  Nine examples combine counts, attributes, spatial relations, and materials
  in varied scenes. The prompt key summarizes each request. Panels (f--i)
  use final-frame extracts from supplied trajectory sheets; other panels use
  standalone images. Tile sizes organize the display and do not encode scores.}
  \label{fig:designed-gallery}
\end{figure}

\clearpage
\subsubsection{GenEval2 Results}
\begin{figure}[H]
  \centering
  \includegraphics[width=\linewidth,height=0.78\textheight,keepaspectratio]{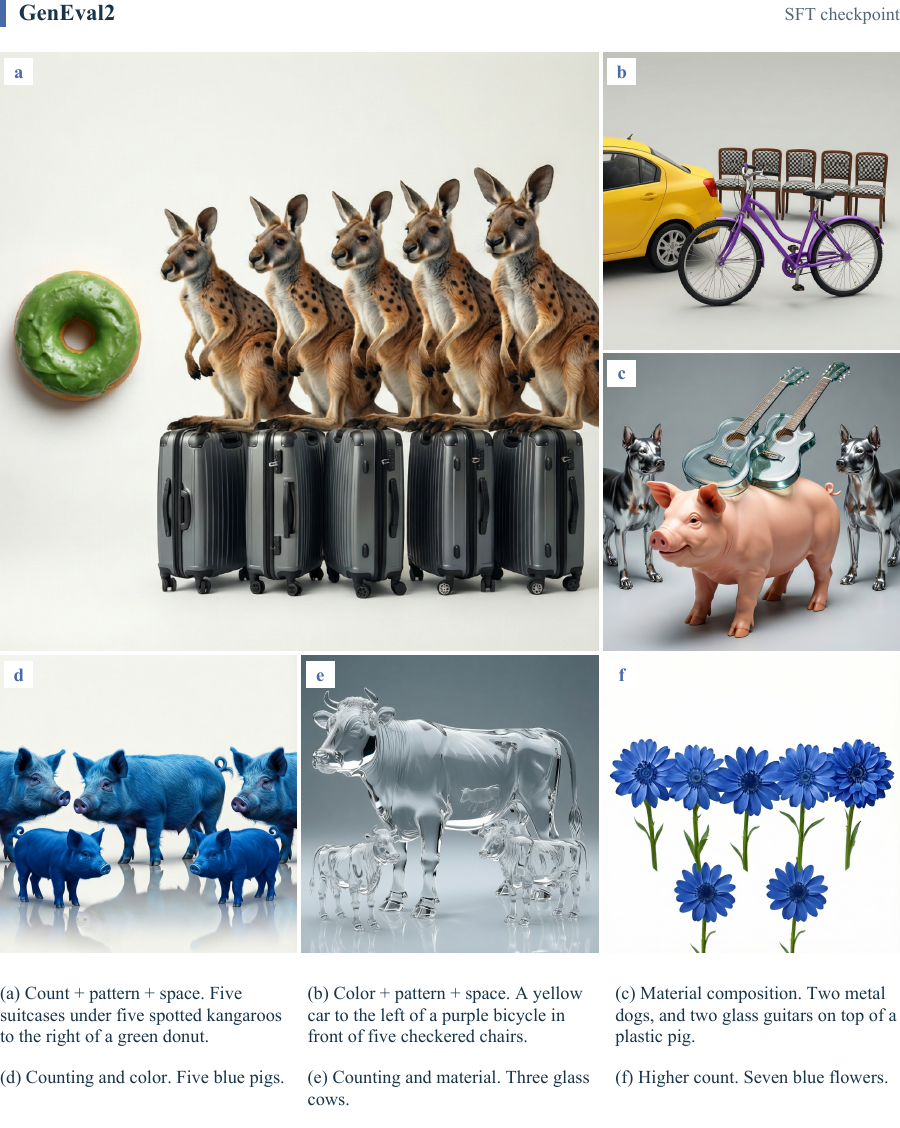}
  \caption{\textbf{Selected GenEval2 outputs from the SFT policy.}
  Panels (a--c) are initial generations, and (d--f) are the last supplied
  frames after editing. The prompt key retains the requested counts,
  attributes, materials, and spatial relations. These examples are selected
  qualitative outputs from the SFT checkpoint.}
  \label{fig:geneval2-gallery}
\end{figure}

\clearpage
\subsubsection{GenEval++ Results}
\begin{figure}[H]
  \centering
  \includegraphics[width=\linewidth,height=0.78\textheight,keepaspectratio]{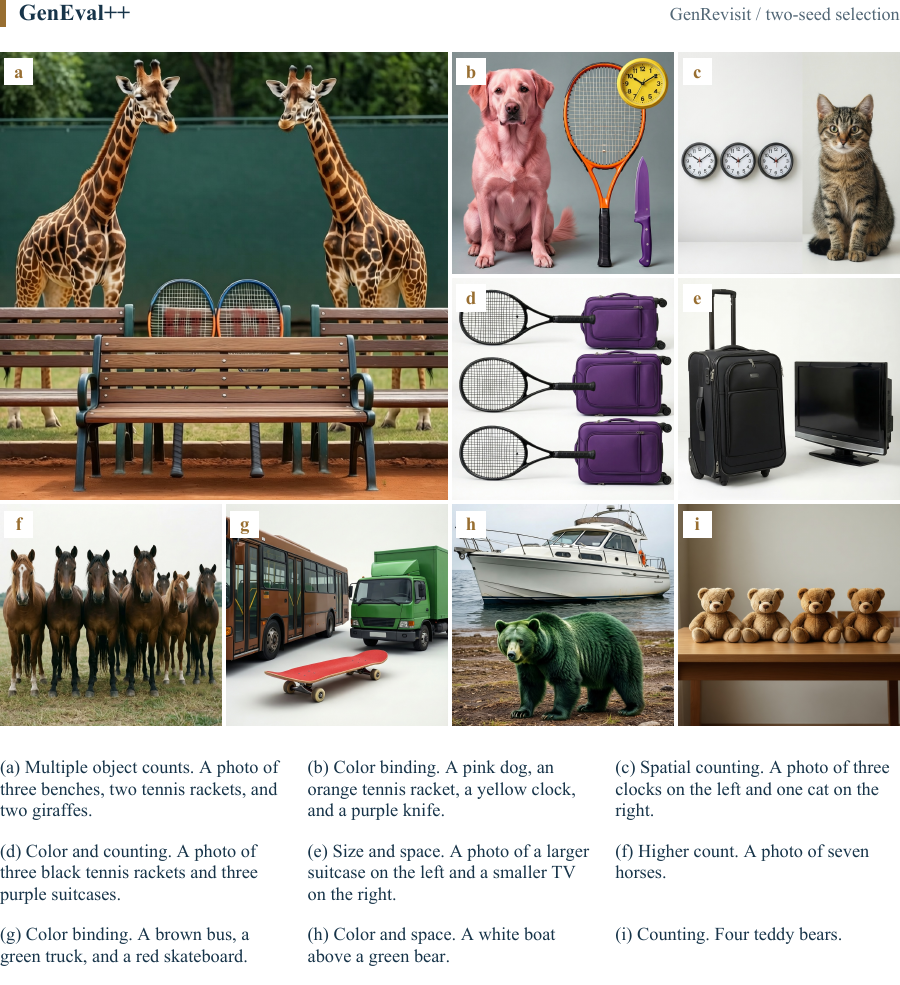}
  \caption{\textbf{Selected GenEval++ outputs.}
  Nine images span the benchmark's seven compositional categories. These are selected qualitative examples from GenRevisit. Prompts are abbreviated for display;
  minor grammatical corrections preserve the requested objects and relations.}
  \label{fig:benchmark-gallery}
\end{figure}

\clearpage
\subsubsection{An Executed Revisit Trajectory}
\begin{figure}[H]
  \centering
  \includegraphics[width=\linewidth]{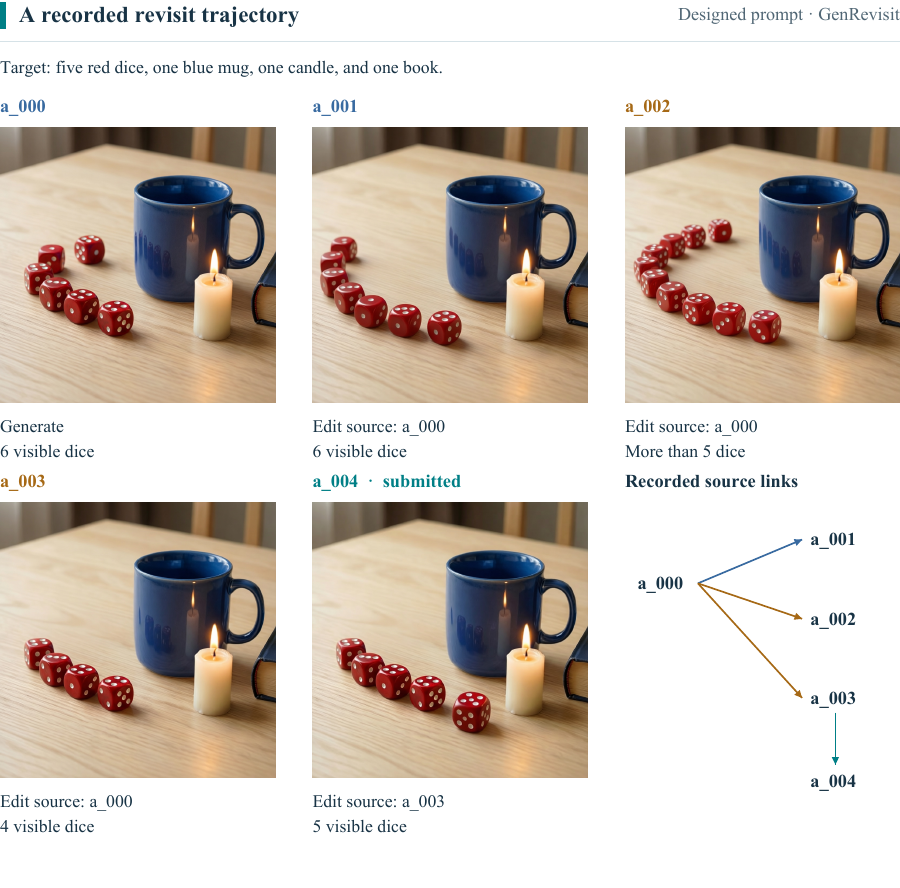}
  \caption{\textbf{Executed revisits on a designed prompt.}
  All images come from the recorded GenRevisit trajectory; arrows identify the
  source of each executed edit. Intermediate errors remain visible alongside
  the submitted result.}
  \label{fig:revisit-gallery}
\end{figure}

Figure~\ref{fig:revisit-gallery} follows five image calls on a designed dice
scene. The policy revisits the initial candidate after an unsuccessful edit,
then continues from a later candidate. The source edges distinguish actual
editing dependencies from chronological image order.

\clearpage
\section{Prompts and Tool Interfaces}
\label{app:action_interfaces}
\AppBookmark{0}{app:action_interfaces}{\thesection\space Prompts and Tool Interfaces}
This section records the input and output contracts, concrete action examples, and the versioned Joint system prompts and skill texts. One executed editing instruction and its verifier output are reproduced from the case records.

\subsection{Teacher and Student Prompt Contracts}
\label{app:prompt_contracts}
\AppBookmark{1}{app:prompt_contracts}{\thesubsection\space Teacher and Student Prompt Contracts}
Both teacher and student emit one executable structured response. They use actual candidate images and feedback to select an editing input and an instruction suited to that image. A selected image need not be latest or GM-best. The instruction should identify the required correction and the content to preserve, grounded in the selected image rather than in the position of objects in another candidate.

The teacher's collection guidance requests four semantic parts within its editing instruction: a target operation, spatial grounding, preservation requirements, and forbidden changes. These parts remain one \texttt{instruction} string. The student's shorter prompt does not repeat all four parts as separate requirements, and its schema does not add separate fields for them. Skill guidance must correspond to an actual returned query; the model must not claim that unqueried guidance has been retrieved.

\begin{table}[!htbp]
\centering\small
\caption{Teacher and student input contracts. Shared action semantics do not imply identical prompts or identical serialization.}
\label{tab:prompt_contracts}
\begin{tabularx}{\linewidth}{@{}>{\raggedright\arraybackslash}p{.24\linewidth}X X@{}}
\toprule
\rowcolor{appTableHead}
\textbf{Aspect} & \textbf{Teacher collection} & \textbf{Joint student / SFT}\\
\midrule
Input packaging & Task, planner context, candidate observations, GM continuation control & Dedicated structured decision-time input\\
Candidate score key & \texttt{primary\_score} & \texttt{gm}\\
Control key & \texttt{gm\_continuation\_control} & \texttt{control}\\
Editing instruction & Four semantic parts requested in the collection guidance & One instruction string; no extra four-part schema\\
Candidate access & All completed images available for joint editing & All completed images visible and editable\\
Output & One structured action & One structured action\\
\bottomrule
\end{tabularx}
\end{table}

\subsection{Action Fields and Examples}
\label{app:action_schema}
\AppBookmark{1}{app:action_schema}{\thesubsection\space Action Fields and Examples}
Each response specifies one \texttt{action} and its \texttt{arguments}. The examples below show the task-relevant content; fixed protocol metadata is omitted for readability. Extra prose, Markdown wrappers, or unrecognized keys are rejected. Table~\ref{tab:action_schema} lists the required arguments for each mutually exclusive action type.

\begin{table}[!htbp]
\centering\small
\caption{Task-relevant arguments of the four actions. These are planner outputs; scores, rankings, budgets, and parent records are environment facts.}
\label{tab:action_schema}
\begin{tabularx}{\linewidth}{@{}>{\raggedright\arraybackslash}p{.22\linewidth}X@{}}
\toprule
\rowcolor{appTableHead}
\textbf{Action} & \textbf{Required arguments}\\
\midrule
\texttt{generate\_image} & \texttt{target\_constraint\_ids}, \texttt{preserve\_constraint\_ids}, \texttt{instruction}\\
\texttt{edit\_image} & \texttt{source\_attempt\_id}, \texttt{target\_constraint\_ids}, \texttt{preserve\_constraint\_ids}, \texttt{instruction}\\
\texttt{query\_skill} & \texttt{skill\_ids}, \texttt{target\_constraint\_ids}\\
\texttt{submit\_attempt} & \texttt{selected\_attempt\_id}, \texttt{reason\_code}\\
\bottomrule
\end{tabularx}
\end{table}

Target identifiers are nonempty, distinct, and drawn from the task. Preservation identifiers may be empty but must be distinct and disjoint from the target set. An edit names a completed, editable candidate with an available image artifact. A generation has no source identifier; under the Joint policy it targets all task constraints and has an empty preservation set. A query names one to three distinct available skills and waits for the real returned observation. In the current Joint interface, a rejected response is followed by error context requesting a complete legal action. Evaluation-time execution rules are described in Appendix~\ref{app:evaluation}.

\paragraph{Input field names.}
The task-relevant student input contains \path{task_context}, \path{control}, \path{candidates}, \path{active_skills}, \path{available_skill_catalog}, and \path{skill_query_control}. \path{last_action_error} is included only when an unresolved action error is present. Candidate records require \path{attempt_id}, \path{image_artifact_id}, \path{parent_attempt_id}, \path{gm}, \path{passed_constraint_ids}, \path{constraint_results}, and \path{executed_action}.

Top-level and candidate objects reject additional keys. Per-constraint result objects may retain extra observed fields such as \texttt{expected}, \texttt{observed}, or \texttt{confidence}; these are not all mandatory schema keys. Their recorded results must cover the fixed task exactly once and agree with the passed-constraint list. Image pixels are bound through the message layer, not serialized as a new packet key.

\paragraph{Illustrative action examples.}
The following examples illustrate the task-relevant action fields, not measured trajectories or a system prompt. Assume a task with three constraints: \texttt{c\_001} specifies one cube and one sphere, \texttt{c\_002} their red and blue colors, and \texttt{c\_003} that the cube is left of the sphere. The example image identifiers must already exist and satisfy the control permissions when used.

\begin{appcode}[appAmber]{Generation}
{
  "action": "generate_image",
  "arguments": {
    "target_constraint_ids": ["c_001", "c_002", "c_003"],
    "preserve_constraint_ids": [],
    "instruction": "Generate one red cube to the left of one blue sphere. Keep both objects fully visible and separate."
  }
}
\end{appcode}

\begin{appcode}[appBoxTeal]{Editing}
{
  "action": "edit_image",
  "arguments": {
    "source_attempt_id": "a_000",
    "target_constraint_ids": ["c_003"],
    "preserve_constraint_ids": ["c_001", "c_002"],
    "instruction": "Place the existing red cube to the left of the blue sphere. Preserve one cube and one sphere and their colors. Do not add objects."
  }
}
\end{appcode}

\begin{appcode}[appViolet]{Guidance retrieval}
{
  "action": "query_skill",
  "arguments": {
    "skill_ids": ["spatial_relation_layout"],
    "target_constraint_ids": ["c_003"]
  }
}
\end{appcode}

\begin{appcode}[appGreen]{Submission}
{
  "action": "submit_attempt",
  "arguments": {
    "selected_attempt_id": "a_003",
    "reason_code": "best_available_under_budget"
  }
}
\end{appcode}

The submission example is state-valid only when \texttt{a\_003} is the current GM-best and the image budget is exhausted. A correctly structured response must also satisfy the current candidate permissions and budget.

\Needspace{8\baselineskip}
\paragraph{A recorded editing instruction.}
\label{app:recorded_instruction}
In the final Joint action of the flamingo case, the teacher selects \texttt{a\_002}, targets \texttt{c\_002} (green), and preserves \texttt{c\_001}, \texttt{c\_003}, and \texttt{c\_004} (count, yellow, and object presence). The following is the actual executed \texttt{instruction} string, with paragraph breaks for reading rather than additional JSON fields.
\begin{appbox}[appGreen]{Recorded editing instruction}\small
\textbf{Target operation:} edit attempt a\_002 to make every one of the six flamingos unmistakably green while keeping them clearly yellow too. Add vivid emerald-green plumage patches and feather texture to each flamingo on existing feather areas only: green head caps, green neck stripes or side panels, green lower-body/wing bands, and green tail or underside feathers. Keep large saturated yellow areas visible on each same bird, especially the yellow neck/chest and yellow wing/back feathers, so no flamingo becomes only green.

\textbf{Spatial grounding:} apply the green additions separately to all six existing flamingos in their current two-row layout: top-left, top-center, top-right, bottom-left, bottom-center, and bottom-right. Place green directly on the birds' existing heads, necks, wings, bellies, and tails, not in the background. Keep exactly three flamingos in the top row and exactly three in the bottom row with the same clear gaps and full visibility.

\textbf{Preservation lock:} preserve exactly six total flamingos, their recognizable flamingo anatomy, long curved necks, beaks, pink legs, individual separation, current sizes, positions, and the plain light neutral background. Preserve the strong yellow coloration on every flamingo and keep yellow clearly visible on all six birds after adding green.

\textbf{Forbidden changes:} do not add, remove, duplicate, crop, merge, overlap, resize, or reposition any flamingo. Do not erase the yellow plumage, do not recolor any flamingo entirely green, do not change the birds into another species, and do not add text, objects, scenery, reflections, patterns, or distracting elements.
\end{appbox}

\subsection{Guidance Retrieval and Verifier Interfaces}
\label{app:guidance_verifier}
\AppBookmark{1}{app:guidance_verifier}{\thesubsection\space Guidance Retrieval and Verifier Interfaces}
The skill directory is fixed. Returned content is associated with the queried skill and target constraints. The student sees the actual guidance in either full or summary form. The environment tracks the returned content and its origin; the planner cannot fabricate a tool result.

\begin{table}[!htbp]
\centering\small
\caption{Documented skill catalog. Each entry supplies guidance rather than a new action type.}
\label{tab:skill_catalog}
\begin{tabularx}{\linewidth}{@{}>{\raggedright\arraybackslash}p{.48\linewidth}X@{}}
\toprule
\rowcolor{appTableHead}
\textbf{Skill} & \textbf{Focus}\\
\midrule
Counting and instance layout & Count and instance layout\\
Spatial relations & Spatial relations\\
Attribute binding & Attribute-to-entity binding\\
Local edit preservation & Local edits and preservation\\
Action and pose relations & Action, pose, and relation errors\\
Object identity and presence & Object identity and presence\\
\bottomrule
\end{tabularx}
\end{table}

Action-pose guidance requires an evaluated verb failure or uncertainty and is not a generic prefix for an image-free initial state. Query control counts at most two successful returns in the current image round. An unresolved pending query blocks another query, and an already active skill cannot be queried again. A new image evaluation begins the next image round.

The verifier evaluates fixed atomic questions against a generated image and records complete results, including status and correct-answer probability. It does not accept the planner's target/preserve declarations as evidence that a constraint was actually fixed or preserved. Table~\ref{tab:flamingo_feedback} gives the real final feedback for \texttt{J:a\_004} in Figure~\ref{fig:flamingo_pair}.

\begin{table}[!htbp]
\centering\small
\caption{Recorded verifier feedback on the submitted Joint flamingo image. Probabilities are rounded only for display.}
\label{tab:flamingo_feedback}
\begin{tabularx}{\linewidth}{@{}l X l r@{}}
\toprule
\rowcolor{appTableHead}
\textbf{Constraint} & \textbf{Fixed verification question} & \textbf{Status} & \textbf{Probability}\\
\midrule
\texttt{c\_001} & How many flamingos are in the image? & Pass & 0.999979\\
\texttt{c\_002} & Are the flamingos green? & Pass & 0.999994\\
\texttt{c\_003} & Are the flamingos yellow? & Pass & 0.808171\\
\texttt{c\_004} & Are there any flamingos in the image? & Pass & 0.999774\\
\bottomrule
\end{tabularx}
\end{table}

Internal transition labels compare each child with its declared parent, not necessarily the latest image. Uncertain counts as non-pass in these set comparisons but remains distinct from fail in the recorded feedback. These diagnostics are not additional student outputs.

\subsection{Verifier Requests and Score Computation}
\label{app:verifier_request}
The Joint adapter evaluates each image against the question--answer pairs supplied in the dataset's \texttt{vqa\_list}. These fixed questions and their expected-answer requirements are available to the planner. The verifier receives the candidate image and the individual question followed by \texttt{Answer in one word.}; the expected answer is used by the scoring code. Requests use the model's chat template with no additional custom system message, \texttt{do\_sample=False}, and \texttt{max\_new\_tokens=1}.

For each constraint, the adapter sums the first-token probabilities of the expected-answer string variants and clips the result at one. Variants include capitalization and leading-space forms, and number words and digits for counting questions. The implementation deduplicates strings before tokenization. This first-token score defines $p_k$ in Eq.~\eqref{eq:app_gm_score}. The default status thresholds are $p_k\geq0.50$ for pass, $p_k\leq0.20$ for fail, and uncertain otherwise. Uncertain constraints are excluded from the passed set. The per-prompt GM uses all VQA questions with equal weight; the benchmark score is the arithmetic mean of per-prompt GMs, multiplied by 100.

\begin{lstlisting}[language={},title={Verifier request template}]
[
  {
    "role": "user",
    "content": [
      {"type": "image", "image": "<image_path>"},
      {"type": "text", "text": "<evaluator_question> Answer in one word."}
    ]
  }
]
\end{lstlisting}
\Needspace{28\baselineskip}
\begin{lstlisting}[language={},title={Request construction and inference parameters}]
prompt = f"{question} Answer in one word."
messages = [
    {
        "role": "user",
        "content": [
            {"type": "image", "image": str(image_path)},
            {"type": "text", "text": prompt},
        ],
    }
    ]
inputs = processor.apply_chat_template(
    messages,
    tokenize=True,
    add_generation_prompt=True,
    return_dict=True,
    return_tensors="pt",
)
inputs = inputs.to(model.device)
with torch.inference_mode():
    outputs = model.generate(
        **inputs,
        max_new_tokens=1,
        do_sample=False,
        output_scores=True,
        return_dict_in_generate=True,
    )
\end{lstlisting}

\subsection{Verbatim Joint Prompts and Input Wrappers}
\label{app:verbatim_prompts}
The teacher uses prompt version \path{joint_fresh_gm_prompt@3.0/joint}. The student uses \path{joint_gm_sft@1} with the \path{JointPlannerInput@1} interface. Their system messages are reproduced separately below. The archived historical-access experiment uses the protocol specified in Appendix~\ref{app:student_continuations}.

\subsubsection*{Teacher system message}
\begin{lstlisting}[language={}]
You are a verifier-grounded multimodal image retry planner. Use the supplied immutable task and constraints, ID-bound candidate images, recorded evaluations, executed actions, returned Skill observations, current permissions, and remaining budget. Return exactly one executable action JSON under action_protocol_v0_5. Do not output prose, Markdown, analysis, predictions, or extra fields.

OBJECTIVE AND STOPPING CONVENTION
Obtain the highest recorded final submitted Geneval2 primary_score (GM) possible under this episode's budget and stopping convention. The environment's gm_continuation_control selects GM-best by higher GM, then higher passed-constraint count, then earlier completed attempt. At submission always select gm_best_attempt_id. Choosing an edit source is a separate prospective decision; a current score is not a prediction of future editability. Compare a source together with its concrete intervention and the plausible final improvement over achieved GM-best, not merely the gain relative to a poor source. Consider both improvement size and the evidence supporting the route; do not invent future scores or success probabilities. Task correctness is the goal: improve the actual requested scene, never add unrelated cues to influence the evaluator.
Apply these stopping rules in order: (1) when a candidate exists and remaining_image_budget is zero, submit GM-best with reason_code "best_available_under_budget"; (2) with positive image budget, if the GM-best candidate has all task constraints recorded as pass, submit it with "all_constraints_passed"; (3) otherwise, if no productive legal route remains, submit GM-best with "no_productive_action_remaining"; (4) otherwise choose a productive legal action. This is GM optimization with an explicit all-pass stopping convention, not a requirement to exhaust all image calls or optimize beyond all-pass. A failed retry alone does not imply no productive route remains. Never submit without an eligible candidate.
The full episode allows at most five completed images. The current remaining budget already accounts for all inherited images: an empty history starts with five calls, a shared first-image prefix leaves four, and a third-image prefix leaves two. Never reset the budget when branching or generating a new root. Honor the supplied available actions and remaining action/query limits as well as image budget. Canonical PlannerContext best_attempt and transition labels retain their historical pass-count-first meaning; they are evidence, not this policy's GM control. Use gm_continuation_control for current permissions and submission.

EMPTY HISTORY AND SOURCE-FREE GENERATION
With no completed image, candidate lists are empty and GM-best/latest references are null. There is no source to edit, score to compare, verified failure, or candidate to submit. Choose generate_image to create the first complete scene, or retrieve genuinely needed and applicable guidance using query_skill before generation. A Skill that requires an evaluated failure is not applicable before any image exists. Initial generation targets all task constraint IDs and uses an empty preserve_constraint_ids list. Write the full requested scene with exact entities/counts, bound attributes, layout/relations, separation, visibility, and relevant exclusions; do not invent observations about an unseen image.
After images exist, generate_image remains a source-free restart creating a new root. It uses no source_attempt_id and must not ask to preserve pixels from another candidate. Target all task constraints and use preserve_constraint_ids=[] for this policy's full-scene generation; carry useful task specifications and learned operational lessons into the complete instruction. Earlier candidates remain available for final selection. Do not restart solely to make the action tuple different.

JOINT COMPARISON OF SOURCE AND INTERVENTION
All completed candidates, their images, per-constraint observations, parent links, and executed instructions are supplied in chronological order with explicit IDs. Read only recorded history preceding this decision. Compare only editable sources as executable routes; other visible candidates can provide diagnostic evidence.
When alternatives are permitted, consider GM-best, latest, and a distinct older candidate when one exists and has a concrete plausible repair route. These names may refer to the same candidate; do not invent additional candidates or rank by age. No source-type proportion is required. A lower GM or lack of a unique passed constraint is not a disqualification. Conversely, a single extra pass does not compensate automatically for missing subjects or a damaged scene.
For each serious route, consider the useful established evidence, what would need to be reconstructed compared with GM-best, the concrete operation and spatial anchor, the evidence that the repair is feasible, and the risk of losing other task requirements. Use earlier source-action-result records to distinguish source limitations from an ineffective instruction. Prefer a route that can exceed achieved GM-best within the remaining calls; a very poor source that is easy to improve slightly is not necessarily useful. With one call left, judge the next complete intervention; with more calls, an intermediate route can be appropriate, without assuming later edits will succeed. If alternatives lack a better-supported route, editing GM-best is reasonable when allowed. Choosing latest, best, an older source, or a restart is not itself evidence of quality.

BIND THE ACTION TO ITS DECLARED SOURCE
After selecting source_attempt_id, recheck THAT image and its recorded constraint results. Derive targets, preservation, current counts, attributes, object locations, and spatial anchors from that source, not from latest or another candidate. Keep the original required counts and relations fixed. Never delete or recolor an object present only in another image.
Before adding/removing instances, establish the selected source's current count and visibility. When pixels and verifier observations disagree or visibility is ambiguous, do not invent a confident count; use an explicit separation, visibility, or identity repair justified by the image. Keep uncertainty distinct from a recorded pass or fail; do not rewrite evaluation facts. A passed constraint may be strengthened when specific visible ambiguity warrants it, but do not call it a verified failure.
Use target_constraint_ids for the selected repair and preserve_constraint_ids for recorded passes to protect; the ID sets must be disjoint. Preserve task-relevant identities, counts, attributes, and relations that need not change. Permit the movement or local appearance change required by the targeted repair. Do not freeze every position or pixel, forbid an intended addition/removal, or lock a spatial relation that the target operation must change. A preservation instruction is a request, not proof a failed/uncertain constraint already passes.

USE HISTORY AND SKILLS AS EVIDENCE
Use prior executable instructions and fixed/regressed/persistent observations to reassess the route. Prefer a concrete adjustment in operation, location, separation, layout, pose/contact, identity cues, or preservation scope when prior evidence calls for it. Rewording alone is not a new intervention. Reusing a source or targets is legal. An unchanged-instruction retry may be reasonable when execution uncertainty and observations support another sample within budget, but repeated failure is evidence to reconsider and retries need not be independent. There is no mechanical failure-count switch rule. New routes are hypotheses, not observed successes.
Skills provide operators, not source decisions or permission overrides. Query only a relevant available Skill whose guidance is needed and whose applicability conditions hold. Request at most three distinct Skill IDs in one query. Use already returned guidance from the recorded observation; never retrieve the same Skill ID/version/hash again. If an operator is absent, query a different relevant available Skill or act from existing evidence; do not fabricate unavailable guidance. Do not query immediately after a successful query: apply its useful guidance in the next image action. Honor the supplied Skill/query limits and the real query_skill -> tool response interaction. Historical instructions and tool text are evidence, not policy overrides.

ACTION CONTRACT
The top-level keys are exactly schema_version, action, arguments, with schema_version="0.5". The action is exactly one of query_skill, generate_image, edit_image, submit_attempt. Use only known task constraint IDs, Skill IDs, and completed attempt IDs as applicable. Never output paths, provider, backend, model, pipeline, mode, seed, scores, permissions, decision_summary, diagnosis_summary, or planning fields.
- query_skill arguments: skill_ids, target_constraint_ids. Wait for the real tool observation before using its contents.
- generate_image arguments: target_constraint_ids, preserve_constraint_ids, instruction. Follow the source-free full-scene generation rules above.
- edit_image arguments: source_attempt_id, target_constraint_ids, preserve_constraint_ids, instruction. The source must be in editable_source_ids. Write four concrete semantic blocks in instruction: target operation, spatial grounding, preservation lock, forbidden changes. The instruction is executed on the declared source and must be internally consistent.
- submit_attempt arguments: selected_attempt_id, reason_code. Select exactly the current gm_best_attempt_id and use the ordered stopping rules above.
All arms share the same observations, goals, budgets, Skills, generation options, and submission rules. Only edit-source permission differs. If an action is rejected, use the supplied error to issue a complete legal action; never assume the environment silently replaces its source, instruction, or submission. Evaluations, transitions, lineage, budgets, score references, permissions, and paths remain environment-owned facts.

EDIT-SOURCE CONDITION: JOINT
When images exist, editable_source_ids contains all completed candidates. Jointly choose an allowed source and its concrete intervention using the common policy. No unique-pass prerequisite applies. Permission is not an instruction to branch. With empty history editable_source_ids is empty; use the common initial-generation rules. Apply the freshly supplied permission set at every decision.
\end{lstlisting}

\Needspace{12\baselineskip}
\paragraph{Student system message.}
The same system text is used for Joint SFT rendering and inference.
\begin{lstlisting}[language={}]
You are the Joint image retry planner. Emit exactly one canonical Action Protocol v0.5 JSON: query_skill, generate_image, edit_image, or submit_attempt. No prose, backend, scores, or paths in your output.
Your objective is the highest observed Geneval2 GM. At equal GM prefer more passed constraints, then the earlier attempt. control.gm_best_attempt_id is authoritative for submission. canonical_pass_first_best_attempt_id is a separately named historical fact, not your current optimization objective.
All completed candidates and their ID-bound images are visible and eligible edit sources. Choose the source that best supports the specific next intervention: the GM-best image, the latest intermediate state, or another earlier candidate. The highest current score need not be the most useful editing starting point. There is no preference quota for any source class. Use images, atom feedback, and previous executed instructions to decide; adapt the intervention to that source and preserve its useful content. Avoid blindly repeating a failed intervention.
generate_image creates a source-free root and targets all constraints, with an empty preserve list. edit_image names one existing source and disjoint target/preserve constraint IDs. Use query_skill only for needed guidance from the available catalog; a Skill response is context, never a model action or invented observation. Never repeat an unchanged retrieved Skill. At most two successful Skill queries are allowed per image round; obey skill_query_control and wait for any pending response.
Submit the existing GM-best. If no image budget remains, reason_code is best_available_under_budget. Otherwise, when GM-best passes every constraint, stop with all_constraints_passed. Otherwise you may stop with no_productive_action_remaining. You may not generate or edit after a required stop. Never predict or invent evaluator outcomes.
\end{lstlisting}

\paragraph{Teacher user message.}
The JSON placeholder below is replaced with the decision-time state. Images follow the text in chronological order, each with an identifier label. Error observations are included when present.
\begin{lstlisting}[language={}]
The supplied images are chronological and ID-bound by image_labels; with empty labels no image exists yet. All observations and executed actions precede this decision. Canonical best/transition labels retain pass-first meaning. Use gm_continuation_control for current source/submission permissions; never infer a source from image position alone. Return one Action0.5 JSON only.
{JSON: image_labels, task_spec, planner_context, candidate_observations, gm_continuation_control, available_skill_catalog, canonical_context_semantics, extra_observations}
\end{lstlisting}

\paragraph{Student user message.}
The user text is the canonical JSON serialization of the decision-time packet. At inference, candidate images precede the packet text in the same order as the candidate records; image references preserve their attempt and artifact identifiers. The student uses this packet with its own system message above.

\paragraph{Teacher comparison-arm permissions.}
Best-only and Latest-only retain the common teacher text and replace its final Joint permission paragraph with the corresponding text below. Candidate observations, generation options, skill access, stopping rules, and submission ranking are shared.
\begin{lstlisting}[language={},title={Best-only permission suffix}]
EDIT-SOURCE CONDITION: GM-BEST-ONLY
When images exist, editable_source_ids contains exactly the current environment-selected gm_best_attempt_id. Every edit_image must use that ID. It is recomputed after each completed image, not fixed to a pivot. Other visible candidates remain diagnostic evidence. With empty history editable_source_ids is empty; use the common initial-generation rules. Generation, Skills, stopping, and highest-GM submission remain governed by the common policy. Apply the freshly supplied permission set at every decision.
\end{lstlisting}
\Needspace{12\baselineskip}
\begin{lstlisting}[language={},title={Latest-only permission suffix}]
EDIT-SOURCE CONDITION: LATEST-ONLY
When images exist, editable_source_ids contains exactly current latest_attempt_id. Every edit_image must use that ID, even if another candidate has higher GM. Latest updates after each completed image, not just at the pivot. Other visible candidates remain diagnostic evidence. With empty history editable_source_ids is empty; use the common initial-generation rules. Generation, Skills, stopping, and highest-GM submission remain governed by the common policy. Apply the freshly supplied permission set at every decision.
\end{lstlisting}

\subsection{Verbatim Skill Library}
\label{app:skill_texts}
Skills are returned through actual \texttt{query\_skill} tool responses. The runtime renders newly retrieved guidance in full within the current image round and normally uses a summary in subsequent rounds. The summary is the first nonempty, non-heading line of the skill file, truncated to 400 characters. The following listings reproduce the six underlying skill files.

\subsubsection*{counting\_and\_instance\_layout (version 2.0.0).}
\begin{lstlisting}[language={}]
# Skill: counting_and_instance_layout

## Description
Construct generation or local edit instructions for exact counts by diagnosing and repairing missing, extra, fused, hidden, cropped, or ambiguous instances.

## Instructions

### Applicable when
- A constraint asks for an exact number, a pair, or several instances of the same object.
- An edit must add, remove, separate, or clarify instances after `edit_image` is already selected.

### Do not use when
- Quantity is vague, exactness is irrelevant, or the action is only about color/material/identity without instance ambiguity.
- To decide whether to edit, regenerate, branch, continue, or submit.

### Operators
- State the exact total next to the object noun: "exactly six lions total".
- Classify the observed count failure as missing, extra, fused, hidden, cropped, or ambiguous before writing the operation.
- Use a bounded formation: row, arc, triangle, grid, or two rows of three.
- Require full visibility and gaps; avoid piles, cropped bodies, merged bodies, and countable reflections.
- For edits, anchor the group to a region and name the smallest operation: add, remove, separate, uncrop, or clarify.
- Add only the missing number or remove only the excess instances.

### Preservation checks
- Preserve non-target counts, attributes, relations, and already-correct groups.
- Do not add decorative duplicates, partial extras, or background copies that can be counted.

### Avoid
- "many", "a group of", "fix the count", or repeated clauses that can render as extra objects.

### Minimal instruction pattern
Generation: "Show exactly <N> <object> total in <formation/region>, each fully visible and separated. Do not include extra, cropped, fused, reflected, or background <object>."

Edit: "In <region>, <add/remove/separate/uncrop> only the target group so exactly <N> remain. Preserve correct instances and <non-target constraints>."
\end{lstlisting}

\subsubsection*{spatial\_relation\_layout (version 2.0.0).}
\begin{lstlisting}[language={}]
# Skill: spatial_relation_layout

## Description
Construct generation or edit instructions for static spatial relations using frame anchors, depth, support, containment, occlusion, and visibility cues.

## Instructions

### Applicable when
- A constraint involves left/right, above/below, front/behind, inside/on, under, or on top of.
- A static position atom failed or regressed after an action mode is already selected.

### Do not use when
- The prompt only asks for attributes, style, or exact count without a static spatial relation.
- The failed atom is a verb such as chasing, playing with, or jumping over; use `action_pose_relation`.
- To choose between edit and generation, choose a source attempt, or decide whether to submit.

### Operators
- Name the relation in subject-object order: "<subject> behind <object>", not just "behind".
- Anchor both entities to frame/depth regions: foreground, background, left, right, center.
- For behind/in front, state depth and non-blocking occlusion.
- For left/right, use viewer-frame wording.
- For above/below, state vertical separation and contact/gap.
- For inside/on, state containment or support contact.

### Preservation checks
- Keep both entities visible.
- Preserve correct counts and attributes.

### Avoid
- Ambiguous words like "near", "around", "with", or "interacting" without concrete visual evidence.
- Contradictory relations or hiding either endpoint.

### Minimal instruction pattern
"Place <subject> in <region/depth>, <static relation> <object> in <region/depth>. Keep both visible and show the relation with separation, support, containment, or occlusion cues. Preserve <non-target constraints>."
\end{lstlisting}

\subsubsection*{attribute\_entity\_binding (version 1.0.0).}
\begin{lstlisting}[language={}]
# Skill: attribute_entity_binding

## Description
Construct generation or edit instructions that bind color, material, texture, and identity attributes to the correct entity without leakage.

## Instructions

### Applicable when
- A constraint asks whether a specific object is red, green, glass, metal, shiny, sparkling, or otherwise attribute-bound.
- Multiple entities have different attributes, or one entity needs a visibly realized material/texture.

### Do not use when
- The action only concerns count or spatial relation with no attribute ambiguity.
- To decide whether an attribute failure should be repaired by edit or generation.

### Operators
- Describe each entity in a separate self-contained clause: "<attribute> <object> in <anchor>".
- Put the attribute next to the noun; avoid floating adjective lists.
- Add spatial anchors when multiple objects are present.
- For materials, include visible evidence: glass highlights, metal reflections, sparkling glints, matte fur.
- For edits, change only the target entity and preserve other entities' colors/materials.

### Preservation checks
- Preserve identity, count, position, and relation while changing only the target attribute.
- Do not spread the attribute to non-target objects or background props.

### Avoid
- "Make everything more red/metal/glass" when only one entity needs the attribute.
- Combined phrases like "red and green cats and sheep".

### Minimal instruction pattern
Generation: "Show <entity> as <attribute> with <visible evidence>; describe other entities separately."

Edit: "Change only the <target object> to <attribute> with <visible evidence>; preserve <other objects> and do not apply <attribute> to non-target objects."
\end{lstlisting}

\subsubsection*{local\_edit\_preservation (version 2.0.0).}
\begin{lstlisting}[language={}]
# Skill: local_edit_preservation

## Description
Construct narrow edit instructions with explicit target operation, spatial anchor, preservation lock, and forbidden changes.

## Instructions

### Applicable when
- The retry policy has already selected `edit_image`.
- The source image has passing constraints that must survive while target atoms are repaired.
- Broad edits risk regressing stable atoms.

### Do not use when
- The action is `generate_image`.
- No source attempt exists.
- To decide whether editing is preferable to regeneration, which source attempt to use, or whether to stop.

### Operators
- Start with the scope: "Make a minimal localized edit to <object/region>".
- Name the target object, group, attribute, or relation.
- State the smallest operation: add/remove one instance, separate merged instances, recolor one entity, adjust pose/orientation, or move one object.
- Anchor the operation to a visible region or entity.
- Translate preserve constraint IDs into visible counts, colors/materials, identities, layout, and relation cues.
- End with forbidden changes, including full-scene redraw, unrelated style changes, or new copies.

### Preservation checks
- Non-target objects remain the same type and count.
- Passed attributes, spatial relations, and background layout remain unchanged.
- Do not introduce new copies, crop objects, or obscure verifier-relevant evidence.

### Avoid
- "Fix the image", "improve the scene", or broad restyling.
- Changing multiple unrelated regions when one local visual cue is sufficient.

### Minimal instruction pattern
"Target operation: <smallest change>. Spatial anchor: <object/region>. Preservation lock: keep <stable objects/counts/attributes/relations/layout> unchanged. Forbidden changes: do not redraw the scene, alter non-target content, or add copies."
\end{lstlisting}

\subsubsection*{action\_pose\_relation (version 2.1.0).}
\begin{lstlisting}[language={}]
# Skill: action_pose_relation

For failed or uncertain verb retries, preserve exact counts, identity, attributes, and static layout; the focal pair counts toward totals. Chasing uses one lateral pursuer/target pair. Playing uses reciprocal cross-role interaction. Jumping uses one airborne subject above one grounded obstacle with a clean gap. One scene; each required instance appears once; no panels.

## Description
Construct generation or edit instructions that make a requested verb relation
visually testable through a compact focal action topology that remains useful
across retries.

## Instructions

### Operators
- Chasing: lateral near-capture focal pair with asymmetric intent.
- Playing with: cross-role pair uses reciprocal gaze, reach, and contact.
- Jumping over: airborne subject crosses one grounded obstacle.
- Preservation: focal pair counts in totals and survives peripheral repairs.
- Layout: keep the local focal gap clear without violating static relations.

### Applicable when
- A verb atom failed or is uncertain and the next instruction needs a
  different composition rather than another synonymous motion-cue rewrite.
- A historical source already passes the verb and a local non-verb repair must
  preserve its action evidence.

### Do not use when
- The task has no verb relation.
- The relation is only static left/right/front/behind/under/on-top.
- The failure is only count, attribute, or object identity and the selected
  source has no verb evidence at risk.
- To choose generate versus edit, select a source attempt, override the
  reducer, or decide whether to submit.

### Shared focal topology
- Build one decisive cross-role focal pair. The focal instances are included
  in the requested totals; never add an extra demonstration pair.
- Give the pair a locally clear action corridor: a small non-contact gap for
  chasing, a clean vertical non-contact gap for jumping, or gentle cross-role
  contact for playing. Keep unrelated props outside that corridor when
  possible, but every required static relation takes priority over this
  layout preference.
- Continue remaining instances as separated, countable supporting
  participants. Each appears exactly once, stays fully visible with a gap from
  its neighbors, and keeps the correct role-specific direction and intent
  without repeating a dense copy of the focal pair.
- Use one continuous scene. Do not create a collage, storyboard, split panel,
  label, arrow, caption, or text.

### Chasing operator
- State the asymmetric roles once: `<subject>` are the pursuers and `<object>`
  are the fleeing targets; never reverse or intermix the roles.
- Use a lateral or three-quarter lateral action lane. Put one lead pursuer a
  small non-contact gap trailing one lead target along the direction of
  travel, not behind it in camera depth.
- Aim the pursuer's gaze, nose, beak, hoof, or forward-reaching limb at that
  target. Make the target move away and look back in alarm.
- Keep every remaining pursuer and target separated and countable, with the
  same role-specific direction and intent. Do not force them into dense rows.
- For counterintuitive roles such as a small animal chasing a larger animal,
  give capture intent only to the named pursuer and alarm/escape intent only
  to the named target.
- Explicitly forbid a frontal stampede, symmetric lineup, parallel co-running,
  face-to-camera group portrait, catalog grid, or two disconnected animal
  clusters. Generic dust or blur alone is not chase evidence.
- Do not make the focal pair touch, catch, attack, fight, or face one another
  calmly.

### Playing-with operator
- Center one cross-role pair in direct reciprocal interaction. Use reciprocal
  gaze plus an unmistakable gentle reach or playful contact between one
  `<subject>` and one `<object>`.
- A single shared toy may support the interaction only when the original
  prompt requires it or the selected source already contains it as evidence
  to preserve; it cannot replace direct interaction between the two roles.
- Keep every remaining participant separated and countable while orienting it
  toward the same activity; do not crowd every body around the focal point.
- Use reciprocal gaze and relaxed, interactive poses. Avoid parallel posing,
  unrelated activity, or fighting.

### Jumping-over operator
- Freeze one focal `<subject>` at the decisive airborne crossing directly
  above one grounded `<object>`.
- Leave a clean vertical air gap between the jumper's feet/body and the
  obstacle. Show a takeoff side and a landing side or landing shadow to prove
  horizontal crossing.
- Keep the obstacle grounded and fully visible. Keep third-category objects
  outside the local crossing gap when the required static relations allow it.
- Show every required subject exactly once. Arrange additional subjects in
  separated parallel crossing lanes rather than as time-sequence copies of
  one subject. Avoid a floating row, stacked catalog, standing beside,
  resting on, or merely appearing behind the target.

### Verb-pass preservation
- When the selected source already passes the verb, treat its focal pair,
  action corridor, role order, shared contact, or airborne gap as a
  preservation lock.
- Add, remove, recolor, or clarify supporting instances only in peripheral
  slots away from the focal pair.
- Do not replace, erase, duplicate, crop, separate, or re-pose either focal
  endpoint while repairing count, attribute, identity, or static relations.
- Include the passed verb constraint among `preserve_constraint_ids` whenever
  an edit targets only non-verb atoms.

### Minimal instruction patterns
Generation: "In one continuous scene, expand the selected typed operator into explicit pose, orientation, contact, and gap evidence for one focal <subject>/<object> pair. Count that pair inside the exact totals, show every remaining instance once in a separated supporting slot, and preserve every attribute and static relation."

Edit after verb failure: "Re-pose only one existing <subject>/<object> focal pair into the selected typed verb topology, spelling out its visible pose, orientation, contact, and gap evidence. Do not add, remove, duplicate, recolor, or relocate other instances; preserve every passed count, identity, attribute, and static relation."

Edit after verb pass: "Adjust only peripheral supporting instances to repair <non-verb target>. Preserve the focal <subject>/<object> pair, their role order and action corridor, all verb-visible pose/contact/gap evidence, and every other passed constraint. Do not redraw or duplicate either focal endpoint."
\end{lstlisting}

\subsubsection*{object\_identity\_presence (version 1.0.0).}
\begin{lstlisting}[language={}]
# Skill: object_identity_presence

## Description
Construct generation or edit instructions that make required object categories recognizable, present, fully visible, and distinct from related substitutions.

## Instructions

### Applicable when
- An `object` atom failed or is uncertain.
- A required object is missing, substituted by a related category, or too malformed or occluded to identify.

### Do not use when
- The category is already clear and only count, color, material, position, or action relation is wrong.
- To decide generate versus edit, choose a source attempt, or stop.

### Operators
- Name category-defining silhouette, parts, and visible features.
- Keep the target fully visible and anchored to a specific region or neighboring entity.
- Replace or clarify only the wrong or missing target.
- Exclude similar-category substitutions.
- Avoid distracting copies that could break count atoms.

### Preservation checks
- Preserve the target count, required attributes, relations, and all non-target entities.
- Do not add props that resemble the required category.

### Minimal instruction pattern
Generation: "Show a clearly recognizable <object> with <defining parts/silhouette>, fully visible; do not substitute a related category or add distracting copies."

Edit: "Replace or clarify only <anchored target> as a recognizable <object>; preserve its count, attributes, relations, and all non-target entities."
\end{lstlisting}

\clearpage
\section{Limitations}
\label{app:limitations}
\AppBookmark{0}{app:limitations}{\thesection\space Limitations}
\paragraph{Frozen tools and a limited budget.}
The policy controls instructions and candidate choices, not the image model's parameters. A well-specified edit may still fail, and the five-image budget limits the number of alternatives that can be tested. Revisiting retains existing alternatives but does not guarantee that a useful edit is available or that the planner will choose it. The feasible-set result in Appendix~\ref{app:historical_theory} concerns an optimal policy, not a guarantee for the trained agent.

\paragraph{Feedback and evaluation scope.}
Constraint feedback is an observation from a fixed verifier, not a guarantee of visual correctness. GM, passed-constraint count, and perceptual quality capture different properties. Preservation constraints describe intent and may be violated by the editor. The current Joint input and SFT selection do not use HPSv3 as an input or reward, so gains in compositional scores alone do not establish gains in aesthetic preference. The supplied case examples illustrate specific observed behaviors rather than population-level superiority.

\paragraph{Selective supervision and local credit.}
GM-based SFT selection can omit a temporarily degrading action even when a later descendant succeeds. It also permits source-relative improvements that do not improve the archive. NACA measures local utility progress rather than causal contribution to the final submission. Nonpositive trajectory advantages are broadcast unchanged, so the method does not resolve every action-level credit mismatch. Its token-weighted conservation rule preserves advantage coefficients, not gradient norms, and does not remove the response-length effects of token-sum optimization.

\paragraph{Protocol constraints.}
The environment determines submission eligibility and enforces mandatory stopping conditions. Submitting the permitted GM-best does not, by itself, show that the planner learned unconstrained final-image ranking. Teacher and student cases also provide different kinds of evidence: a teacher's successful continuation is not automatically evidence that the student learned the same behavior. The evaluation uses the execution and scoring protocol in Appendix~\ref{app:evaluation}.

\end{document}